\documentclass{article}

    \PassOptionsToPackage{numbers, compress, sort}{natbib}

    \usepackage[final]{neurips_2024}

\usepackage{bbm}
\usepackage{amsmath}
\usepackage{amssymb}
\usepackage[utf8]{inputenc} %
\usepackage[T1]{fontenc}    %
\usepackage{hyperref}       %
\usepackage{url}            %
\usepackage{booktabs}       %
\usepackage{amsfonts}       %
\usepackage{nicefrac}       %
\usepackage{microtype}      %
\usepackage{xcolor}         %
\usepackage{graphicx}
\usepackage{soul}   %
\usepackage{xcolor}  %
\usepackage{subcaption}
\usepackage[heightadjust=all]{floatrow}
\usepackage{adjustbox}
\usepackage{colortbl}
\usepackage{color} 
\usepackage{tikz}
\usetikzlibrary{positioning}
\usepackage{algorithm}
\usepackage{algorithmicx}
\usepackage[noend]{algpseudocode}
\usepackage{bm}
\usepackage{listings}
\usepackage{amsmath}
\usepackage[para,multiple]{footmisc} 
\usepackage{lipsum}
\usepackage{subfig}

\DeclareMathOperator*{\argmin}{arg\,min}

\algnewcommand{\lst}{\texttt{lst}}
\algnewcommand{\slst}{\texttt{slst}}
\algnewcommand{\SEND}{\textbf{send}}

\newcommand*{\ours}{SceneDiffuser}

\newcommand{\boldparagraph}[1]{\textbf{#1}~}
\newcommand{\squad}{\:\:\:}

\definecolor{forestgreen}{rgb}{0.13, 0.55, 0.13}

\colorlet{tableheadcolor}{gray!25} %
\colorlet{tablerowcolor}{gray!10} %
\newcommand{\rowcolorize}{\rowcolor{tablerowcolor}} %

\colorlet{tablebluerowcolor}{blue!10} %
\newcommand{\bluerowcolorize}{\rowcolor{tablebluerowcolor}} %

\definecolor{codegreen}{rgb}{0,0.6,0}
\definecolor{codegray}{rgb}{0.5,0.5,0.5}
\definecolor{codepink}{RGB}{252, 142, 172}
\definecolor{codepurple}{rgb}{0.58,0,0.82}
\definecolor{backcolour}{RGB}{245,245,245}
\lstdefinestyle{mystyle}{
    backgroundcolor=\color{backcolour},   
    commentstyle=\color{magenta},
    keywordstyle=\color{blue},
    numberstyle=\tiny\color{codegray},
    stringstyle=\color{codepurple},
    basicstyle=\fontfamily{\ttdefault}\footnotesize,
    breakatwhitespace=false,         
    breaklines=true,                 
    keepspaces=true,    
    frame=single,
    numbersep=5pt,                  
    showspaces=false,                
    showstringspaces=false,
    showtabs=false,                  
    tabsize=2,
    classoffset=1, %
    keywordstyle=\color{violet},
    classoffset=0,
}
\lstdefinestyle{prompt}{
    language=sh,
    basicstyle=\footnotesize\ttfamily\color{white},
    keywordstyle=\color{white},
    emphstyle=\color{black},
    commentstyle=\color{green},
    showstringspaces=false,
    numbers=left,
    numberstyle=\footnotesize\color{white},
    frame=single,
    framesep=5pt,
    rulecolor=\color{black},
    xleftmargin=15pt,
    framexleftmargin=10pt,
    backgroundcolor=\color{black},
    moredelim=[il][\textcolor{black}]{$ $},
    moredelim=[is][\textcolor{black}]{\%\%}{\%\%},
}

\definecolor{waymogreen}{rgb}{0,0.91,0.62}

\definecolor{waymoblue}{rgb}{0,0.47,1}

\newcommand{\Tau}{\mathcal{T}}

\title{
\ours{}: Efficient and Controllable Driving Simulation Initialization and Rollout
}

\author{Chiyu Max Jiang \quad Yijing Bai\thanks{Equal contribution core technical contributors (alphabetically ordered).} \quad Andre Cornman$^{*}$ \quad Christopher Davis$^{*}$ \quad Xiukun Huang$^{*}$ \\
\textbf{Hong Jeon}$^{*}$ \squad \textbf{Sakshum Kulshrestha}$^{*}$ \squad \textbf{John Lambert}$^{*}$ \squad \textbf{Shuangyu Li}$^{*}$ \squad \textbf{Xuanyu Zhou}$^{*}$ \\
\textbf{Carlos Fuertes} \quad \textbf{Chang Yuan} \quad \textbf{Mingxing Tan} \quad \textbf{Yin Zhou} \quad \textbf{Dragomir Anguelov} \\
~ \\
Waymo LLC \\
}

\begin{document}
\setlength{\textfloatsep}{10pt plus 1.0pt minus 2.0pt}
\setlength{\floatsep}{12pt plus 1.0pt minus 1.0pt}
\newfloatcommand{capbtabbox}{table}[][\FBwidth]
\newfloatcommand{capbalgbox}{algorithm}[][\FBwidth]

\maketitle

\begin{abstract}
Realistic and interactive scene simulation is a key prerequisite for autonomous vehicle (AV) development. In this work, we present \ours{}, a scene-level diffusion prior designed for traffic simulation. It offers a unified framework that addresses two key stages of simulation: scene initialization, which involves generating initial traffic layouts, and scene rollout, which encompasses the closed-loop simulation of agent behaviors. While diffusion models have been proven effective in learning realistic and multimodal agent distributions, several challenges remain, including controllability, maintaining realism in closed-loop simulations, and ensuring inference efficiency. To address these issues, we introduce amortized diffusion for simulation. This novel diffusion denoising paradigm amortizes the computational cost of denoising over future simulation steps, significantly reducing the cost per rollout step (16x less inference steps) while also mitigating closed-loop errors. We further enhance controllability through the introduction of generalized hard constraints, a simple yet effective inference-time constraint mechanism, as well as language-based constrained scene generation via few-shot prompting of a large language model (LLM). Our investigations into model scaling reveal that increased computational resources significantly improve overall simulation realism.
We demonstrate the effectiveness of our approach on the Waymo Open Sim Agents Challenge, achieving top open-loop performance and the best closed-loop performance among diffusion models.
\end{abstract}

\section{Introduction}
\begin{figure}
\centering
\input{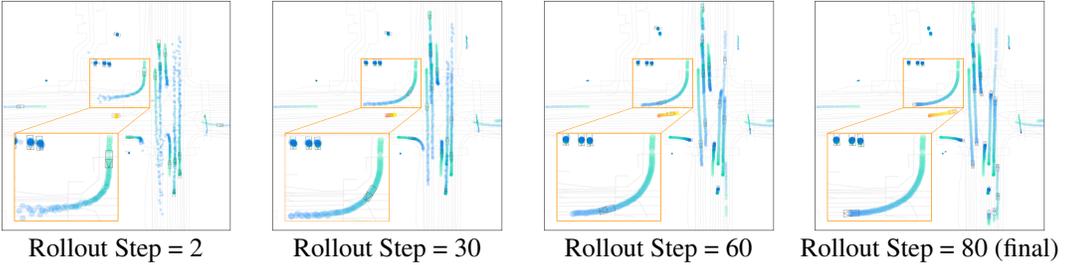}
\vspace{-5mm}
\vspace{-.5em}
\caption{\ours{}: a generative prior for simulation initialization via log perturbation, agent injection, and synthetic scene generation, and for efficient closed-loop simulation at 10Hz via amortized diffusion. It progressively refines initial trajectories throughout the rollout. Environment sim agents are in \textcolor{waymogreen}{green}-\textcolor{waymoblue}{blue} gradient (temporal progression), AV agent in \textcolor{orange}{orange}-\textcolor{yellow}{yellow}, and synthetic agents in \textcolor{red}{red}-\textcolor{purple}{purple}.}
\label{fig:teaser}
\end{figure}

\vspace{-.5em}
Simulation environments allow efficient and safe evaluation of autonomous driving systems \cite{Bergamini21icra_SimNet, Tan21cvpr_SceneGen,Feng22arxiv_TrafficGen, Pronovost23neurips_ScenarioDiffusion,Mahjourian24icra_UniGen, Wang23tr_MultiverseTransformer,Igl22icra_Symphony,Philion24iclr_Trajeglish,Zhang23icra_TrafficBots,Xu22arxiv_BITS,Vinitsky22neurips_Nocturne}. Simulation involves initialization (determining starting conditions for agents) and rollout (simulating agent behavior over time), typically treated as separate problems \cite{Sun24ral_DriveSceneGen}. Inspired by diffusion models' success in generative media, such as video generation \cite{Gupta23arxiv_WALT,Brooks24_SoraVideoGenModels} and video editing (inpainting \cite{Lugmayr22cvpr_RePaint,Nitzan24arxiv_LazyDiffusionImageEditing,Mu24arxiv_EditableImageElements}, extension, uncropping etc.), we propose \ours{}, a unified spatiotemporal diffusion model that addresses both initialization and rollout for autonomous driving, trained end-to-end on logged driving scenes. To our knowledge, \ours{} is the first model to jointly enable scene generation, controllable editing, and efficient learned closed-loop rollout (Fig.~\ref{fig:teaser}).

One challenge in simulation is evaluating long-tail safety-critical scenarios \cite{Bergamini21icra_SimNet, Tan21cvpr_SceneGen,Feng22arxiv_TrafficGen, Pronovost23neurips_ScenarioDiffusion,Mahjourian24icra_UniGen}. While data mining can help, such scenarios are often rare. We address this by learning a generative scene realism prior that allows editing logged scenes or generating diverse scenarios. Our model supports scene perturbation (modifying a scene while retaining similarity) and agent injection (adding agents to create challenging scenarios). We also enable synthetic scene generation on roadgraphs with realistic layouts. We design a protocol for specifying scenario constraints, enabling scalable generation, and demonstrate how a few-shot prompted LLM can generate constraints from natural language.

Given a scene, realistically simulating agents and AV behavior is challenging \cite{Wang23tr_MultiverseTransformer,Igl22icra_Symphony,Philion24iclr_Trajeglish,Zhang23icra_TrafficBots,Xu22arxiv_BITS,Vinitsky22neurips_Nocturne}. Unlike motion prediction tasks \cite{Varadarajan2021_MultiPathPP, Sapp2020_Multipath, Nayakanti2022_Wayformer, Shi2022_MTR, Jiang23cvpr_MotionDiffuser, Ngiam22icml_SceneTransformer} where entire future trajectories are jointly predicted in a single inference, simulator predictions are iteratively fed back into the model, requiring realism at each step. This poses challenges: distributional drift from compounding errors, high computational cost for models like diffusion, and the need to simulate various perception attributes realistically.

We propose Amortized Diffusion for simulation rollout generation, a novel approach for amortizing the cost of the denoising inference over a span of physical steps that effectively addresses the challenges of simulation realism due to closed-loop drift and inference efficiency. Amortized diffusion iteratively carries over prior predictions and refines them over the course of future physical steps (see Sec.~\ref{sec:scene_rollout} and Fig.~\ref{fig:schematic_amortized}). This allows our model to produce stable, consistent, and realistic simulated trajectories, while requiring only a \textit{single} denoising function evaluation at each physical step while jointly simulating all perception attributes at each step. Experiments show that Amortized Diffusion not only requires 16x less model inferences per step, but is also significantly more realistic.

In summary, \ours{}'s main contributions are:
\begin{itemize}
  \item A unified generative model for scene initialization and rollout, jointly learning distributions for agents, timesteps, and perception features including pose, size and type.
  \item A novel amortized diffusion method for efficient and realistic rollout generation, significantly improving trajectory consistency and reducing closed-loop error.
  \item Controllable scene initialization methods, including log perturbation, agent injection, and synthetic generation with a novel hard constraint framework and LLM.
  \item Investigation of model scaling, showing increased compute effectively improves realism. 
  \item Demonstration of effectiveness on the Waymo Open Sim Agents Challenge, achieving top open-loop performance and the best closed-loop performance among diffusion models.
\end{itemize}

\vspace{-2mm}
\section{Related Work}

\subsection{Data-driven Agent Simulation}
A variety of generative models have been explored for scene initialization and simulation, including autoregressive models \cite{Tan21cvpr_SceneGen,Mahjourian24icra_UniGen,Feng22arxiv_TrafficGen}, cVAEs \cite{Suo21cvpr_TrafficSim}, cGANs \cite{Bergamini21icra_SimNet}, and Gaussian Mixture Models (GMMs) \cite{Feng22arxiv_TrafficGen,Tan23corl_LCTGen}. For closed-loop rollouts, these models have been extended with GMMs \cite{Wang23tr_MultiverseTransformer}, GANs \cite{Igl22icra_Symphony}, AR models over discrete motion vocabularies \cite{Philion24iclr_Trajeglish}, cVAE \cite{Zhang23icra_TrafficBots}, and deterministic policies \cite{Xu22arxiv_BITS,Vinitsky22neurips_Nocturne}. Open-loop rollouts have also been explored using cVAE \cite{Rempe22cvpr_GeneratingAccidentProneScenarios}.

\subsection{Diffusion Models for Agent Simulation}
\boldparagraph{Open-loop Sim} Open-loop simulation generates behavior for agents that all lie within one's control, i.e. does not receive any external inputs between steps. Open-loop simulation thus cannot respond to an external planner stack (AV), the evaluation of which is the purpose of simulation. Diffusion models have recently gained traction in multi-agent simulation, particularly in open-loop scenarios (multi-agent trajectory forecasting) \cite{Seff23iccv_MotionLM, Philion24iclr_Trajeglish}, using either single-shot or autoregressive (AR) generation. Single-shot approaches employ spatiotemporal transformers in ego-centric \cite{Jiang23cvpr_MotionDiffuser, choi2023dice} or scene-centric frames with motion/velocity deltas \cite{Yang24arxiv_WorldCentricDiffusionTransformer, Guo23arxiv_SceneDM}. Soft guidance techniques enhance controllability \cite{Janner22icml_diffuser, Zhong23corl_CTG++}. DJINN \cite{Niedoba23neurips_DiffJointIntNav} uses 2d condition masks for flexible generation.

\boldparagraph{Closed-loop Sim} Closed-loop simulation with diffusion remains challenging due to compounding errors and efficiency concerns. Chang \emph{et al.} \cite{Chang23arxiv_ControllableClosedLoopDiff} explore route and collision avoidance guidance in closed-loop diffusion, while VBD \cite{Huang24arxiv_VBD} combines denoising and behavior prediction losses with a query-centric Transformer encoder \cite{Shi24tpami_MTR++}. VBD found it computationally infeasible to replan at a 1Hz frequency in a receding horizon fashion over the full WOSAC test split due to the high diffusion inference cost, therefore testing in open-loop except over 500 selected scenarios.

\boldparagraph{Initial Condition Generation} Diffusion-based initial condition generation has also been studied \cite{Lu24icra_SceneControl}. Pronovost \emph{et al.} \cite{Pronovost23neurips_ScenarioDiffusion,Pronovost23icraw_GenDrivingScenesDiffusion} adapt the LDM framework to rendered scene images, while SLEDGE \cite{Chitta24arxiv_SLEDGE} and DriveSceneGen \cite{Sun24ral_DriveSceneGen} diffuse initial lane polylines, agent box locations, and AV velocity.

\begin{figure}[t]
    \centering
    \includegraphics[trim={3cm 5.5cm 1.3cm 4.4cm },clip,width=\linewidth]{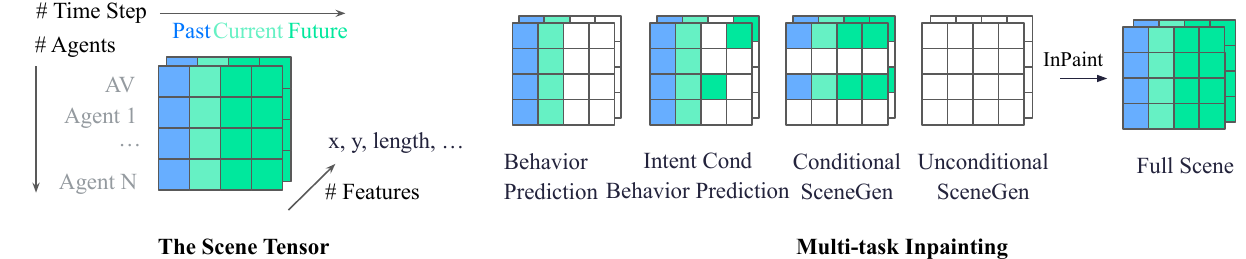}
    \vspace{-1.8em}
    \caption{We formulate various different tasks, including behavior prediction, conditional scenegen and unconditional scenegen as inpainting tasks on the scene tensor. We represent the scene tensor as a normalized tensor $x\in\mathbb{R}^{A\times \Tau\times D}$, for the number of agents, timesteps and feature dimensions.}
    \label{fig:scene-tensor-inpaint}
\end{figure}

\subsection{Diffusion for Temporal World Modeling and Planning}
Outside of the autonomous driving domain, diffusion models have proven effective for world simulation through video and for planning. Various diffusion models for 4d data have been proposed, often involving spatiotemporal convolutions and attention mechanisms \cite{Ho22neurips_VideoDiffusionModels,Ho22_ImagenVideo, Singer23iclr_MakeAVideo}. In robotics, diffusion-based temporal models leverage Model Predictive Control (MPC) for closed-loop control \cite{Chi23rss_DiffusionPolicy} and have shown state-of-the-art performance for imitation learning \cite{Pearce23iclr_ImitatingHumanBehaviourDiffusion}.%

Similar to our Amortized Diffusion approach, TEDi \cite{Zhang23arxiv_TEDi} proposes to entangle the physical timestep and diffusion steps for human animation, thereby reducing $O(T \cdot \Tau)$ complexity for $\Tau$ physical timesteps and $T$ denoising steps to $O(\Tau)$. However, we are the first work to demonstrate the effectiveness of this approach for reducing closed-loop simulation errors, and the first to extend it to a multi-agent simulation setting.

\section{Method}
\subsection{Scene Diffusion Setup}

We denote the scene tensor as $\bm{x}\in\mathbb{R}^{A\times \Tau\times D}$, where $A$ is the number of agents jointly modeled in the scene, $\Tau$ is the total number of modeled physical timesteps, and $D$ is the dimensionality of all the features that are jointly modeled. We learn to predict the following attributes for each agent: positional coordinates $x, y, z$, heading $\gamma$, bounding box dimensions $l, h, w$, and object type $k\sim \{\text{AV, car, pedestrian, cyclist}\}$. We model all tasks considered in \ours{} as multi-task inpainting on this scene tensor. Given an inpainting mask $\bar{\bm{m}}\in\mathbb{B}^{A\times \Tau\times D}$, the corresponding inpainting context values $\bar{\bm{x}}:=\bar{\bm{m}}\odot\bm{x}$, a set of global context $\bm{c}$ (such as roadgraph and traffic signals), and a validity mask for a given agent at a given timestep $\bar{\bm{v}}\in{\mathbb{B}^{A, \Tau}}$ (to account for there being $<A$ agents in the scene or for occlusion), we train a diffusion model to learn the conditional probability $p(\bm{x} | \mathcal{C})$, where $\mathcal{C}:=\{\bar{\bm{m}}, \bar{\bm{x}}, \bm{c}, \bar{\bm{v}}\}$. See Fig.~\ref{fig:scene-tensor-inpaint} for an illustration of the scene tensor.

\boldparagraph{Feature Normalization} To simplify the diffusion model's learning task, we normalize all feature channels before concatenating them along $D$ to form the scene tensor. We first encode the entire scene in a scene-centric coordinate system, namely the AV's coordinate frame just before the simulation commences. We then scale $x, y, z$ by fixed constants, $l, h, w$ by their standard deviation, and one-hot encode $k$. See Appendix~\ref{subsec:normalization} for more details.
This simple yet generalizable process allows us to jointly predict float, boolean, and even categorical attributes by converting into a normalized space of floats.
After generating a scene tensor $\bm{x}$, we apply a reverse process to obtain the generated features.

\begin{figure}[t]%
    \centering
    \includegraphics[trim={0.5cm 9.2cm 2.5cm 0cm},clip,width=\linewidth]{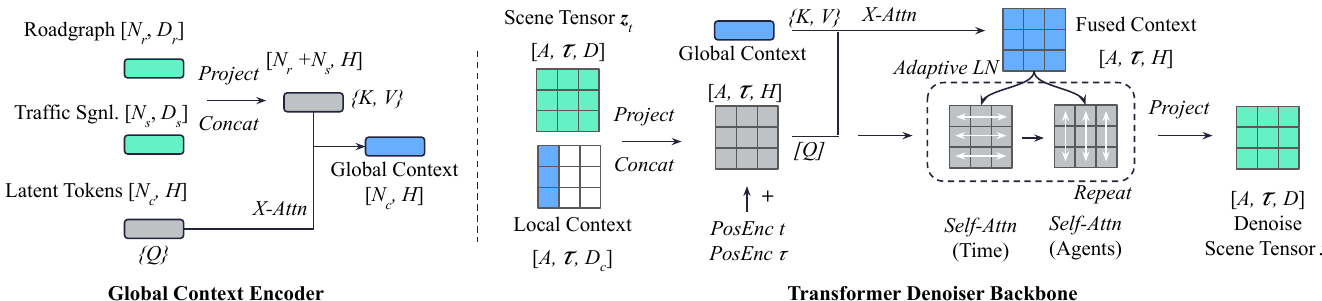}
    \vspace{-1em}
  \caption{\ours{} architecture. Global scene context is encoded into a fixed number of $N_c$ tokens via a Perceiver IO \cite{Jaegle2021_Perceiver} encoder. The noisy scene tokens are fused with local and global context, then used to condition a spatiotemporal transformer-based backbone \cite{Vaswani17nips_AttentionIsAllYouNeed} via Adaptive LayerNorm (AdaLN) \cite{Peebles23iccv_DiT}. Input/output tensor are in \textcolor{waymogreen}{green}, context tensors in \textcolor{waymoblue}{blue}, and ops in \textit{italics}.}
  \label{fig:schematic_architecture}
\end{figure}

\boldparagraph{Diffusion Preliminaries}
We adopt the notation and setup for diffusion models from \cite{Hoogeboom2023_SimpleDiffusion}. The forward diffusion process gradually adds Gaussian noise to $\bm{x}$. The noisy scene tensor at diffusion step $t$ can be expressed as $\mathbf{q}(\bm{z}_t|\bm{x}) = \mathcal{N}{(\bm{z}_t|\alpha_t \bm{x}, \sigma_t^2 \bm{I})}$,
where $\alpha_t$ and $\sigma_t$ are parameters which control the magnitude and variances of the noise schedule under a variance-preserving model. Therefore $\bm{z}_t=\alpha_t \bm{x} + \sigma_t \bm{\epsilon}_t$, where $\bm{\epsilon}_t\sim\mathcal{N}(0, \bm{I})$. One major departure from the classic diffusion setup in our amortized diffusion regime is that we do not assume a uniform noise level $t\in\mathbb{R}$ for the entire scene tensor $\bm{x}$. Instead, we have $t\in \mathbb{R}^{\Tau}$ where $t$ can be relaxed to have a different value per physical timestep in the scene tensor as described in Sec.~\ref{sec:scene_rollout}. We utilize the commonly used $\alpha$-cosine schedule where $\alpha_t=\cos (\pi t/2)$ and $\sigma_t = \sin (\pi t/2)$. At the highest noise level of $t=1$, the forward diffusion process completely destroys the initial scene tensor $\bm{x}$ resulting in $\bm{z}_t = \bm{\epsilon}_t \sim \mathcal{N}{(0,\bm{I})}$. Assuming a Markovian transition process, we have the transition distributions $q(\bm{z}_t|\bm{z}_s) = \mathcal{N}(\bm{z}_t|\alpha_{ts}\bm{z}_s,\sigma_{ts}^2\bm{I})$, where $\alpha_{ts}=\alpha_t/\alpha_s$ and $\sigma_{ts}^2=\sigma_t^2-\alpha_{ts}^2\sigma_s^2$ and $t>s$. In the denoising process, conditioned on a single datapoint $\bm{x}$, the denoising process can be written as
\begin{align}
    q(\bm{z}_s|\bm{z}_t,\bm{x}) = \mathcal{N}(\bm{z}_t|\bm{\mu}_{t\rightarrow s},\sigma_{t\rightarrow s}^2\bm{I}) ,
    \label{eqn:denosing_step}
\end{align}
where $\bm{\mu}_{t\rightarrow s}=\frac{\alpha_{ts}\sigma_s^2}{\sigma_t^2}\bm{z}_t + \frac{\alpha_s \sigma_{ts}^2}{\sigma_t^2}\bm{x}$ and $\sigma_{t\rightarrow s}=\frac{\sigma_{ts}^2\sigma_s^2}{\sigma_t^2}$. In the denoising process, $\bm{x}$ is approximated using a learned denoiser $\hat{\bm{x}}$. Following \cite{Hoogeboom2023_SimpleDiffusion} and \cite{Salimans2022_ProgressiveDistillation}, we adopt the commonly used \textit{v prediction}, defined as $\bm{v}_t(\bm{\epsilon}_t, \bm{x})=\alpha_t \bm{\epsilon}_t - \sigma_{t}\bm{x}$. We trained a model parameterized by $\bm{\theta}$ to predict $\bm{v}_t$ given $\bm{z}_t$ , $t$ and context $\mathcal{C}$: $\hat{\bm{v}}_t := \hat{\bm{v}}_{\theta}(\bm{z}_t, t, \mathcal{C})$. The predicted $\hat{\bm{x}}_t$ can be recovered via $\hat{\bm{x}}_t = \alpha_t \bm{z}_t - \sigma_t \hat{\bm{v}}_t$. The model is end-to-end trained with a single loss:
\begin{align}
    \mathbb{E}_{(\bm{x}, \mathcal{C})\sim \mathcal{D}, t\sim\{\mathcal{U}(0, 1); \hat{\bm{t}}\}, \bm{m}\sim \mathcal{M}, \epsilon_t\sim \mathcal{N}(0, \bm{I})}[||\hat{\bm{v}}_{\theta}(\bm{z}_t, t, \mathcal{C}) - \bm{v}_t(\bm{\epsilon}_t, \bm{x})||_2^2] ,
    \label{eqn:loss_fn}
\end{align}
$\mathcal{D} = \{(\bm{x}_i, \mathcal{C}_i) | i = 1, 2, \cdots, |\mathcal{D}|\}$ is the dataset containing paired agents and scene context data, $t$ is probabilistically either sampled from a uniform distribution, or sampled as a monotonically increasing temporal schedule $\hat{\bm{t}}$, where $\hat{\bm{t}}_{\tau} = \max\big(0, (\tau - \Tau_{\text{history}})/\Tau_{\text{future}}\big)$ to facilitate amortized rollout which will be discussed in Sec.~\ref{sec:scene_rollout}. Each is sampled with 50\% probability. $\mathcal{M} = \{\bar{\bm{m}}_{\text{bp}} \odot \bar{\bm{m}}_{\text{control}}, \bar{\bm{m}}_{\text{scenegen}} \odot \bar{\bm{m}}_{\text{control}}\}$ is the set of inpainting masks for the varied tasks.

\boldparagraph{Scene Diffusion Tasks} Different tasks are fomulated as inpainting problems (Fig.~\ref{fig:scene-tensor-inpaint}).

\emph{Scene Generation (SceneGen)}: Given the full trajectory of some agents, generate the full trajectory of other agents. We have $\bar{\bm{m}}_{\text{scenegen}}\in\mathbb{R}^{A, 1, 1}$ (broadcastable to $\Tau$ timesteps and $D$ features), where $\bar{\bm{m}}_{\text{scenegen, a}}\sim Pr(X=A_{\text{select}}/A_{\text{valid}})$, where $A_{\text{select}}\sim \mathcal{U}(0, A_{\text{valid}})$ is the number of agents sampled to be selected as inpainting conditions out of $A_{\text{valid}}$ valid agents in the scene.

\emph{Behavior Prediction (BP)}: Given past and current data for all agents, predict the future for all agents. We have $\bar{\bm{m}}_{\text{bp}}\in\mathbb{R}^{1, \Tau, 1}$ (broadcastable to $A$ agents and $D$ features), where $\bar{\bm{m}}_{\text{bp}, \tau} = \mathcal{I}(\tau < \Tau_{history})$.

\emph{Conditional SceneGen and Behavior Prediction}: Both scenegen and behavior prediction masks are multiplied by a control mask at training time to enable controllable scenegen and controllable behavior prediction at inference time. We have $\bar{\bm{m}}_{\text{control}}\in\mathbb{R}^{A, \Tau, D}$, where $\bar{\bm{m}}_{\text{control}, (a, \tau, d)} = I_a\cdot I_{\tau} \cdot I_d, I_a\sim Pr(X=A_{\text{control}}/A_{\text{valid}}), I_{\tau}\sim Pr(X=\Tau_{\text{control}}/\Tau), I_d\sim Pr(X=p_d)$ where $p_d$ of the corresponding feature channel. This allows us to condition on certain channels, such as positions $x, y$ with or without specifying other features such as type and heading.

\textbf{Architecture} We present a schematic for the \ours{} architecture in Fig.~\ref{fig:schematic_architecture}, consisting of two end-to-end trained models: a global context encoder and a transformer denosier backbone. Validity $\bar{\bm{v}}$ is used as a transformer attention mask within
the transformer denoiser backbone.

\textbf{Diffusion Sampler} We use DPM++ \cite{Lu22arxiv_dpm++} with a Heun solver. We utilize 16 denoising steps for our one-shot experiments and for our amortized diffusion warmup process. %

\begin{figure}[t!]
    \centering
    \includegraphics[trim={0 7cm 0 2.6cm},clip,width=\linewidth]{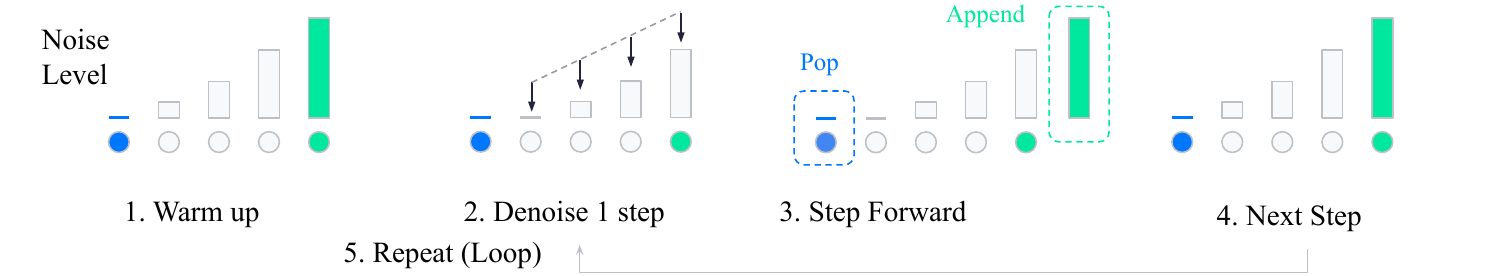}
    \vspace{-6mm}
    \caption{Amortized diffusion rollout procedure. The warm up step initializes the future predictions for the entire future horizon, which is then perturbed by a monotonic noise schedule $\hat{t}$. The trajectory is iteratively denoised by one step at each simulation step.}.
    \label{fig:schematic_amortized}
\end{figure}

\vspace{-2mm}
\subsection{Scene Rollout}
\label{sec:scene_rollout}

Future prediction with no replanning (`One-Shot`) is not used in simulation due to its non-reactivity, and forward scene inference, under the standard diffusion paradigm (`Full AR'), is computationally intensive due to the double for-loop over both physical rollout steps and denoising diffusion steps \cite{Zhang23arxiv_TEDi}. Moreover, executing only the first step while discarding the remainder leads to inconsistent plans that result in compounding errors. We adopt an amortized autoregressive (`Amortized AR`) rollout, aligning the diffusion steps with physical timesteps to amortize diffusion steps over physical time, requiring a single diffusion step at each simulation step while reusing previous plans.

We illustrate the three algorithms in Algorithm~\hyperref[algo:one-shot]{1}-\hyperref[algo:amortized-AR]{3} using the same model trained with a noise mixture $t\sim\{\mathcal{U}(0, 1); \hat{\bm{t}}\}$ (Eqn.~\ref{eqn:loss_fn}). We also illustrate Algorithm~\hyperref[algo:amortized-AR]{3} in Fig.~\ref{fig:schematic_amortized}.  We denote the total number of timesteps $\Tau=H+F$, where $H, F$ denote the number of past and future steps. We denote $\bm{x}:=\bm{x}^{[-H:F]}$ to be the temporal slicing operator where $\bm{x}^{[0]}$ is the final history step.

\boldparagraph{Input:} 
Global context $\mathbf{c}$ (roadgraph and traffic signals), history states $\bm{x}^{[-H:0]}$, validity $\bar{\bm{v}}$.

\boldparagraph{Output:} 
Simulated observations for unobserved futures $\hat{\bm{x}}^{[1:F]}$.
\vspace{-4mm}

\begin{figure}[t!]
\begin{floatrow}\BottomFloatBoxes
\capbalgbox[.47\textwidth]{
    \begin{algorithm}[H]
    \tiny
    \captionsetup{labelformat=empty}
    \caption{ \textbf{Algorithm 1 One-Shot} (Open-Loop)} \label{algo:one-shot}
    \begin{algorithmic}[1]
    \State $ \operatorname{OneShot}(\bm{x}^{[-H:0]}, \mathcal{C})$:
    \State $\bm{z}_{1} \sim \mathcal{N}(0, I)$
    \For{$t=1, \cdots, \Delta t, 0$} \Comment{For each diffusion timestep}
        \State $\hat{\bm{x}}_t \gets \alpha_t \bm{z}_t - \sigma_t \hat{\bm{v}}_{\theta}(\bm{z}_t, t, \mathcal{C})$  \Comment{V-prediction}
        \State $\hat{\bm{x}}_t \gets \bm{x}^{[-H:0]}\odot \bar{\bm{m}}_{\text{bp}} + \hat{\bm{x}}_t \odot (\sim\bar{\bm{m}}_{\text{bp}})$ \Comment{Apply inpainting}
        \State $\bm{z}_s \sim q(\bm{z}_s | \bm{z}_t, \hat{\bm{x}}_t) $
        \State $\bm{z}_t \gets \bm{z}_s$
    \EndFor
    $\hat{\bm{x}} \gets \bm{z}_0$
    \State\Return $\hat{\bm{x}}^{[1:F]}$
    \end{algorithmic}
        \end{algorithm}
}{
    \vspace{-2em}
}
\capbalgbox[.52\textwidth]{
    \begin{algorithm}[H]
    \tiny
    \captionsetup{labelformat=empty}
      \caption{ \textbf{Algorithm 2 Full AR} (Closed-Loop)} \label{algo:naive-AR}
    \begin{algorithmic}[1]
    \State $\operatorname{FullAR}(\bm{x}^{[-H:0]}, \mathcal{C})$:
    \State $\hat{\bm{x}} \gets \bm{x}^{[-H:0]}$
    \For{$\tau=0, ..., \mathcal{T}-1$} \Comment{For each physical timestep} %
    
        \State $\hat{\bm{x}}^{[\tau+1:\tau+F]} \gets \operatorname{OneShot}(\hat{\bm{x}}^{[\tau-H:\tau]}, \mathcal{C})$ \Comment{Update buffer at indices}
    \EndFor
    \State\Return $\hat{\bm{x}}^{[1:F]}$
    \vspace{11.7mm}
    \end{algorithmic}
\end{algorithm}
}{
    \vspace{-2em}
}
\end{floatrow}
\end{figure}

\begin{figure}[h!]
\begin{floatrow}\BottomFloatBoxes
\capbalgbox[.80\textwidth]{
    \begin{algorithm}[H]
    \captionsetup{labelformat=empty}
    \caption{ \textbf{Algorithm 3 Amortized AR} (Closed-Loop)} \label{algo:amortized-AR}
    \tiny
    \begin{algorithmic}[1]
    \State $\operatorname{AmortizedAR}((\bm{x}^{[-H:0]}, \mathcal{C}))$:
    \State $\hat{\bm{x}} \gets \operatorname{OneShot}(\bm{x}^{[-H:0]}, \mathcal{C})$ \Comment{Warm-Up}
    \State $\hat{\bm{x}}^{[0:F]}\gets\alpha_{\hat{\bm{t}}}\hat{\bm{x}}^{[0:F]} + \sigma_{\hat{\bm{t}}}\bm{\epsilon}$ \Comment{Add noise $\hat{\bm{t}}$}
    \For{$\tau=1, ..., \Tau$} \Comment{For each physical timestep}
        \State $\hat{\bm{x}}^{[\tau:\tau+F]} \gets \alpha_{\hat{\bm{t}}} \hat{\bm{x}}^{[\tau:\tau+F]} - \sigma_{\hat{\bm{t}}} \hat{\bm{v}}_{\theta}(\hat{\bm{x}}^{[\tau:\tau+F]}, \hat{\bm{t}}, \mathcal{C})$ \Comment{Recover solution from v-prediction}
        \State $\hat{\bm{x}}^{[\tau:\tau+F]}\gets\alpha_{\hat{\bm{t}}}\hat{\bm{x}}^{[\tau:\tau+F]} + \sigma_{\hat{\bm{t}}}\bm{\epsilon}$ \Comment{Add noise $\hat{\bm{t}}$}
        \State $\hat{\bm{x}}^{[\tau-H:\tau+F]} \gets \hat{\bm{x}}^{[\tau-H:\tau]}\odot \bar{\bm{m}}_{\text{bp}} + \hat{\bm{x}}^{[\tau-H:\tau+F]} \odot (\sim\bar{\bm{m}}_{\text{bp}})$ \Comment{Apply inpainting}
    \EndFor
    \State\Return $\hat{\bm{x}}^{[1:F]}$
    \end{algorithmic}
    \end{algorithm}
}{
}
    
\end{floatrow}
\end{figure}

\setcounter{figure}{4}
\begin{figure}[h!]%
\begin{floatrow}
\ffigbox[.48\textwidth]{%
    \centering
    \vspace{-5mm}
    \includegraphics[trim={1cm 0 0 0},width=\linewidth]{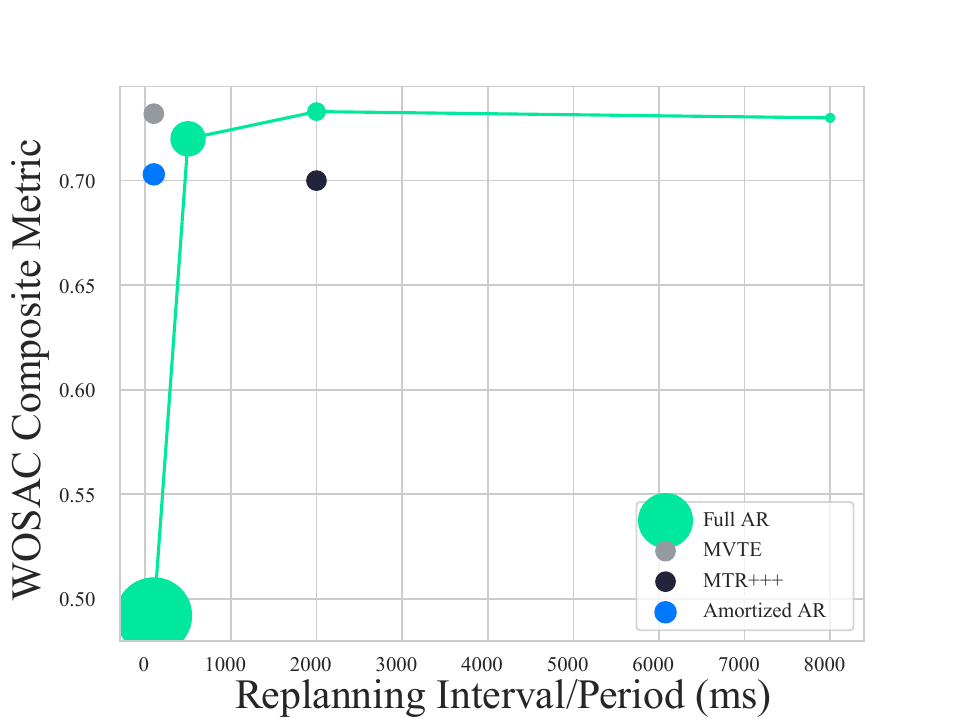}
}{%
  \vspace{-1.5em}
  \caption{We compare the influence of replan rate on performance for our Full AR and Amortized AR models. Circle radius $\propto$ $\#$ inference calls over the simulation. At 10Hz, Amortized AR requires 16x less model inference per step and is more realistic compared to Full AR.}
  \label{fig:full-ar-replan-rate}
}
\ffigbox[.48\textwidth]{%
    \vspace{-5mm}
    \centering
}{%
  \vspace{-1.5em}
  \includegraphics[trim={0 6cm 0 2cm},width=\linewidth]{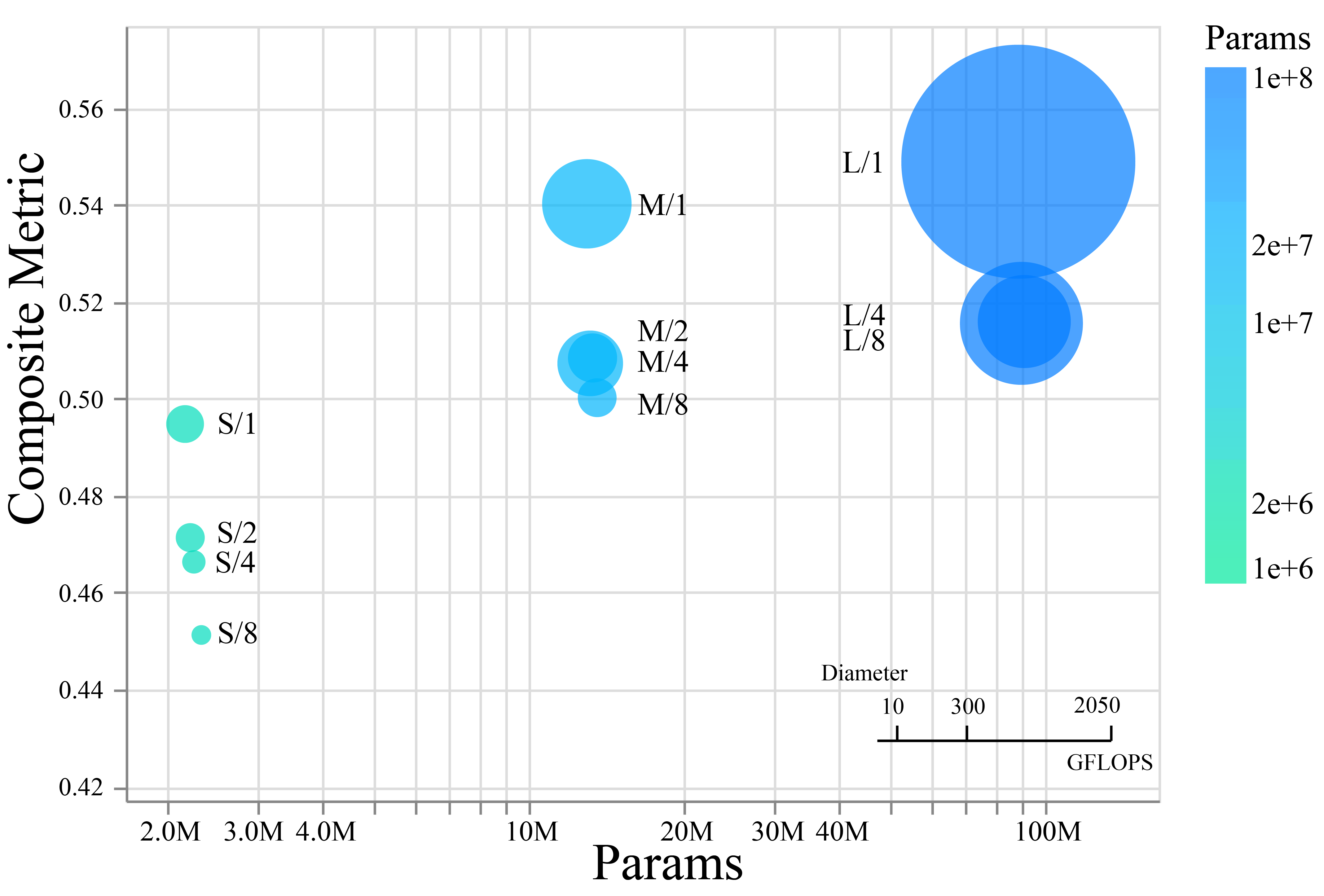}
  \caption{Scene generation realism with model parameter and resolution scaling \protect\footnotemark. Decreased temporal patch sizes (i.e. increased temporal resolution) and increased parameters are both effective for improving realism via compute scaling. Circle radius $\propto$ compute GFLOPs.}
  \label{fig:model-scaling}
}
\end{floatrow}
\end{figure}
\footnotetext{We omit L/2 due to training collapse.}

\subsection{Controllable Scene Generation}

To simulate long-tail scenarios such as rare behavior of other agents, it is important to effectively insert controls into the scene generation process. To do so, we input an inpainting context scene tensor $\bar{\bm{x}}$ , where some pixels are pre-filled. Through pre-filled feature values in $\bar{\bm{x}}$, we can specify a particular agent of a specified type to be appear at a specific position at a specific timestamp.

\boldparagraph{Data Augmentation via Log Perturbation}
The diffusion framework makes it straightforward to produce additional perturbed examples of existing ground truth (log) scenes. Instead of starting from pure noise $\bm{z}_t \sim \mathcal{N}(0, \bm{I})$ and diffusing backwards from $t\rightarrow 0$, we take our original log scene $\bm{x}'$ and add noise to it such that our initial $\bm{z}_t = \alpha_t \bm{x}' + \bm{\epsilon}_t$ where $\bm{\epsilon}_t \sim \mathcal{N}(0, \sigma_t \bm{I})$. Starting the diffusion process at $t=0$ yields the original data, while $t=1$ produces purely synthetic data. For $t\in(0, 1)$, higher values increase diversity and decrease resemblance to the log. See Figs.~\ref{fig:teaser} and \ref{fig:exp_log_perturbation} (Appendix).

\boldparagraph{Language-based Few-shot Scene Generation}
The diffusion model inpaint constraints can be defined through structured data such as a Protocol Buffer\footnote{\url{https://protobuf.dev/}} (`proto'). Protos can be converted into inpainting values, and we leverage the off-the-shelf generalization capabilities of a publicly accessible chat app powered by a large language model (LLM)\footnote{\\The chat app is available at \url{gemini.google.com}, powered by Gemini V1.0 Ultra at the time of access.}, to generate new Scene Diffusion constraints protos solely using natural language via few-shot prompt engineering. We show example results generated by the LLM in Fig.~\ref{fig:exp_scenegen_viz}. Details in the Appendix (\ref{subsec:prompts}).

\vspace{-2mm}
\subsection{Generalized Hard Constraints}
Users of simulation often require agents to have specific behaviors while maintaining realistic trajectories. However, diffusion soft constraints \cite{Zhong23icra_CTG, Zhong23corl_CTG++, Niedoba23neurips_DiffJointIntNav} require a differentiable cost for the constraint and do not guarantee constraint satisfaction. Diffusion hard constraints \cite{Lugmayr22cvpr_RePaint} are modeled as inpainting values and are limited in their expressivity.

Inspired by dynamic thresholding \cite{Saharia2022_Imagen} in the image generation domain, where intermediate images are dynamically clipped to a range at every denoising step, we introduce \emph{generalized hard constraints} (GHC), where a generalized clipping function is iteratively applied at each denoising step. We modify Eqn.~\ref{eqn:denosing_step} such that at each denoising step
$\bm{\mu}_{t\rightarrow s}=\frac{\alpha_{ts}\sigma_s^2}{\sigma_t^2}\bm{z} + \frac{\alpha_s \sigma_{ts}^2}{\sigma_t^2}\text{clip}(\bm{x})$, where $\text{clip}(\cdot)$ denotes the GHC-specific clipping operator. 
See more details on constraints in Appendix~\ref{subsec:appendix_constraint_definitions}.

We qualitatively demonstrate the effect of hard constraints for unconditional scene generation in Fig.~\ref{fig:hard_constraint_clipping}. Applying hard constraints post-diffusion removes overlapping agents but results in unrealistic layouts, while applying the hard constraints after each diffusion step both removes the overlapping agents and takes advantage of the prior to improve the realism of the trajectories. We find that the basis on which the hard constraints operate is important: a good constraint will modify a significant fraction of the scene tensor (for example, shifting an agent's entire trajectory rather than just the overlapping waypoints), or else the model "rejects" the constraint on the next denoising step.

\begin{figure}[t!]%
\begin{floatrow}
\ffigbox[.48\textwidth]{%
    \vspace{-2mm}
    \centering
    \input{tikz/amortized_quality}
    \vspace{-4mm}
}{%
    \caption{Full AR quality deteriorates at increasing replan rates due to compounding errors. Amortized AR retains a high level of realism even at 10 Hz while being more efficient.}
    \label{fig:varied_replan_rate_vs_amortized}
}
\ffigbox[.48\textwidth]{%
    \vspace{-2mm}
    \centering
    \input{tikz/hard_constraints}
    \vspace{-4mm}
}{%
\caption{
Applying no-collision constraints prevents collisions (\textcolor{red}{red}-\textcolor{purple}{purple}) in generated scenes (b, c). Iteratively applying constraints with every diffusion step further enhances realism (c vs b).
}
\label{fig:hard_constraint_clipping}
}
\end{floatrow}
\end{figure}

\begin{figure}[t!]%
\begin{floatrow}
\capbtabbox[.57\textwidth]{%
\begin{adjustbox}{width=\linewidth}
\begingroup
    \begin{tabular}{l | c | c | c | c}
        \toprule
\rowcolorize    \textsc{Metrics} & \textsc{M/1} & \textsc{M/1+GHC} & \textsc{L/1} & \textsc{Log} \\
        \midrule
                \textsc{Composite Metric} & 0.516 & \textbf{0.558} & 0.549 & 0.593 \\
        \midrule
\rowcolorize    \textsc{Linear Speed} & 0.326 & 0.327 & \textbf{0.331} & 0.339 \\
                \textsc{Linear Accel.} & \textbf{0.387} & 0.383 & 0.378 & 0.445 \\
\rowcolorize    \textsc{Angular Speed} & 0.529 & \textbf{0.562} & 0.534 & 0.572 \\
                \textsc{Angular Accel.} & 0.595 & \textbf{0.608} & 0.588 & 0.625 \\
\rowcolorize    \textsc{Dist. to Obj.} & 0.154 & \textbf{0.176} & 0.174 & 0.192 \\
                \textsc{Collision} & 0.692 & \textbf{0.841} & 0.794 & 0.875 \\
\rowcolorize    \textsc{Time to Collision} & 0.826 &  \textbf{0.827} & \textbf{0.827} & 0.842 \\
                \textsc{Dist. to Rd. Edge} & 0.164 & 0.165 & \textbf{0.176} & 0.204 \\
\rowcolorize    \textsc{Offroad} & 0.546 & 0.549 & \textbf{0.566} & 0.605 \\
        \bottomrule
    \end{tabular}
\endgroup
\end{adjustbox}
}{%
  \caption{Realism metrics on WOMD \textit{val.} for scene generation for the M/1 model, M/1 with hard constraints, L/1, and oracle log distribution matching (see Sec.~\ref{sec:scaling}). Realism can be improved through hard constraints or scaling.}%
  \label{tab:scenegen_womd}
}
\ffigbox[.40\textwidth]{%
    \centering
    \includegraphics[width=.85\linewidth]{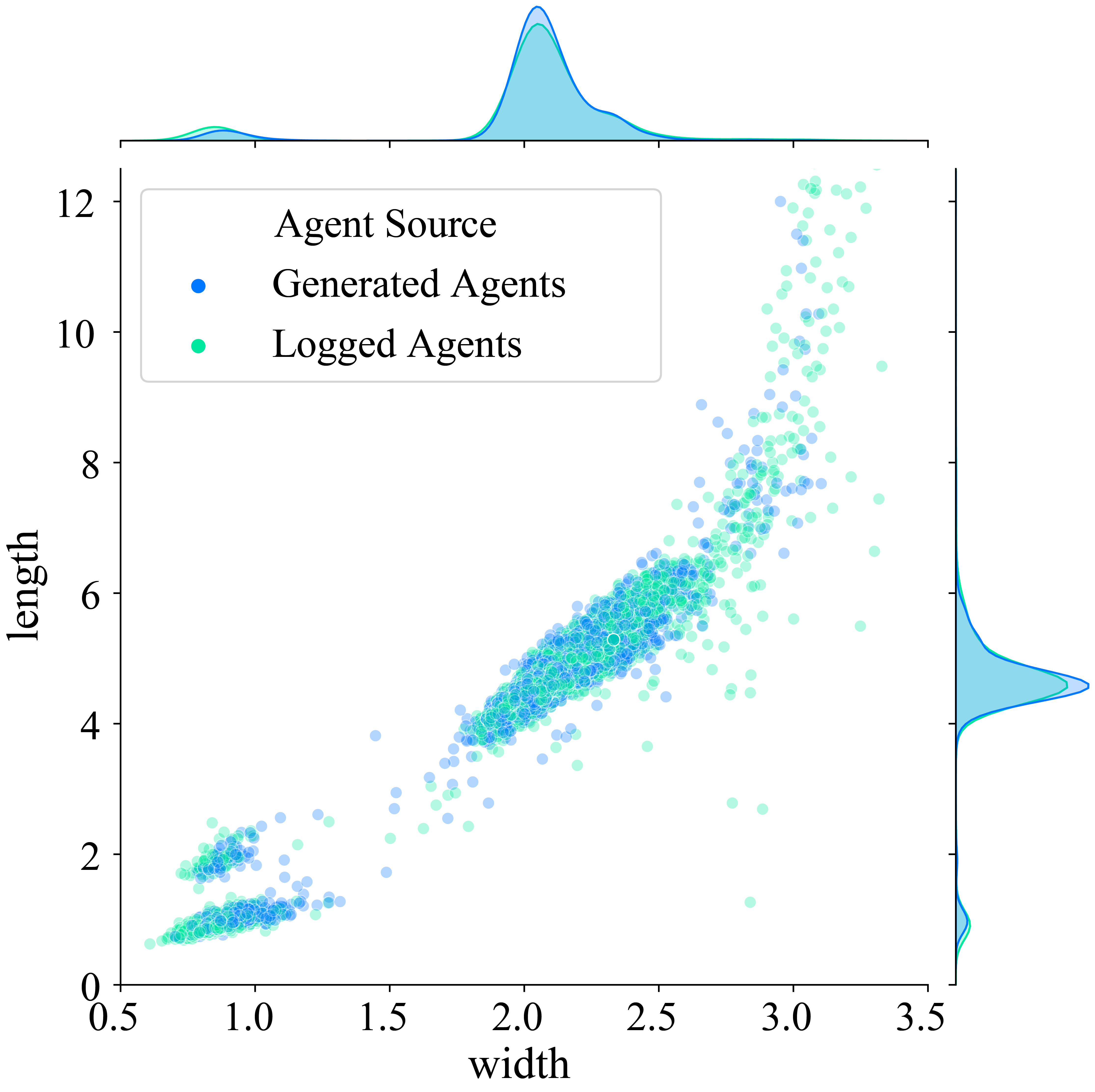}
}{%
  \caption{Generated vs logged distribution. \ours{} learns realistic joint distributions across modeled features such as length and width.}
  \label{fig:scenegen_distribution_comparison}
}
\end{floatrow}
\end{figure}

\vspace{-3mm}
\section{Experimental Results}

\boldparagraph{Dataset} We use the Waymo Open Motion Dataset (WOMD)\cite{Ettinger21iccv_WOMD} for both our scene generation and agent simulation experiments. WOMD includes tracks of all agents and corresponding vectorized maps in each scenario, and offers a large quantity of high-fidelity object behaviors and shapes produced by a state-of-the-art offboard perception system.

\vspace{-2mm}
\subsection{Simulation Rollout}
\boldparagraph{Benchmark} We evaluate our closed-loop simulation models on the Waymo Open Sim Agent Challenge (WOSAC) \cite{Montali23neurips_wosac} metrics (see Appendix \ref{sec:wosac-metrics-details}), a popular sim agent benchmark used in many recent works \cite{Guo23arxiv_SceneDM,Yang24arxiv_WorldCentricDiffusionTransformer, Huang24arxiv_VBD, Philion24iclr_Trajeglish, Wang23tr_MultiverseTransformer}. Challenge submissions consist of x/y/z/$\gamma$ trajectories representing centroid coordinates and heading of the objects’ boxes that must be generated in closed-loop and with factorized AV vs. agent models. WOSAC uses the test data from the Waymo Open Motion Dataset (WOMD)\cite{Ettinger21iccv_WOMD}.  Up to 128 agents (one of which must represent the AV) must be simulated in each scenario for the 8 second future (comprising 80 steps of simulation), producing 32 rollouts per scenario for evaluation. In a small departure from the official setting, we utilize the logged validity mask as input to our transformer and unify the AV and agents' rollout step for simplicity.

\boldparagraph{Evaluation}
In Tab.~\ref{tab:wosac-results}, we show results on WOSAC. We show that Amortized AR (10 Hz) not only requires 16x fewer model inference calls, but is also significantly more realistic than Full AR at a 10Hz replan rate. In Amortized AR, we re-use the plan from the previous step, leading to increased efficiency and consistency. The one-shot inference setting is equivalent to Full AR with no replanning (0.125 Hz) and achieves comparably higher realism, though as it is not executed in closed-loop, it is not reactive to external input in simulation, and thus not a valid WOSAC entry.

In Figs.~\ref{fig:full-ar-replan-rate} and \ref{fig:varied_replan_rate_vs_amortized}, we investigate the effects of varied replan rates to simulation realism. While high replan frequency leads to significant degredation in realism under the Full AR rollout paradigm, Amortized AR significantly reduces error accumulation while being $16\times$ more efficient.

In Tab.~\ref{tab:wosac-per-metric-scores-test}, we compare against the WOSAC leaderboard with the aforementioned modifications. We achieve top open-loop performance and the best closed-loop performance among diffusion models.

\begin{figure}[t!]%
\begin{floatrow}
\capbtabbox[.48\textwidth]{%
    \begin{adjustbox}{width=\linewidth}
    \begingroup
    \begin{tabular}{c|c|c|c}
    \toprule
    \rowcolorize& \textsc{One-Shot} & \textsc{Full AR} & \textsc{Amortized AR} \\
    \rowcolorize& \textsc{} & \textsc{(10 Hz)} & \textsc{(10 Hz)} \\
    \midrule
        \textsc{Composite Score (M/1)} & 0.730 & 0.492 & 0.673  \\ %
        \rowcolorize\textsc{Composite Score (L/1)}  & 0.736 & - & 0.703 \\ %
        \textsc{\# Fn Evals} & $16$ & $16 \cdot 80 = 1280$  & $80 + 16=96$ \\
         \bottomrule
    \end{tabular}
    \endgroup
    \end{adjustbox}
}{%
  \caption{Distrib. realism metrics on WOSAC. L/1 denotes the Large model of patch size 1.}%
 \label{tab:wosac-results}
}
\capbtabbox[.48\textwidth]{%
    \begin{adjustbox}{width=\linewidth}
    \begingroup
    \begin{tabular}{l|c|c|c}
    \toprule
    \rowcolorize& \textsc{Composite Metric} & \textsc{Collision Rate} & \textsc{Offroad Rate} \\
    \rowcolorize& \textsc{$(\uparrow)$} & \textsc{$(\downarrow)$} & \textsc{$(\downarrow)$} \\
    \midrule
        \textsc{-AdaLN-Zero} & -7.99\% & +65.2\% & +29.3\%  \\ %
\rowcolorize        \textsc{-Spatial-Attn} & -14.5\% & +209\% &  +11.8\% \\ %
        \textsc{-MultiTask} & -2.04\% & +39.6\% & +3.24\%  \\ %
\rowcolorize        \textsc{-Size,Type} & 0.68\% & -6.85\% & +2.90\% \\
     \bottomrule
    \end{tabular}
    \endgroup
    \end{adjustbox}
}{%
  \caption{Design analysis and ablation studies.}%
  \label{tab:ablation}
}
\end{floatrow}
\end{figure}

\begin{table*}[t]
    \caption{Per-component WOSAC metric results on the \emph{test} split of WOMD, representing likelihoods. Methods are ranked by composite metric on the 2024 Challenge scores; Closed-loop results within 1\% of the best are in bold (for models with 10Hz replan). Diffusion-based methods marked in \textcolor{waymoblue}{blue}. }
    \centering
    \begin{adjustbox}{width=\linewidth}
    \begingroup
    \begin{tabular}{l|ccccccccccc|cc|cc}
    \toprule
\rowcolorize \textbf{\textsc{Agent Policy}} & \textsc{Replan}	& \textsc{linear}	& \textsc{linear }	& \textsc{ang.}	& \textsc{ang.}	& \textsc{dist.}	& \textsc{collision}	& \textsc{TTC} &	\textsc{dist. to} &	\textsc{offroad} & \textbf{\textsc{Composite}}	& \textsc{ADE}	& \textsc{minADE}	 & \textsc{Collision}  & \textsc{Offroad} \\
\rowcolorize \textsc{} & \textsc{Rate}	& \textsc{speed}	& \textsc{accel.}	& \textsc{speed}	& \textsc{accel.}	& \textsc{to obj.}	& \textsc{}	& \textsc{} &	\textsc{road edge} &	\textsc{} & \textbf{\textsc{Metric}}	& &	 & \textsc{Rate} & \textsc{Rate} \\
\rowcolorize & \textsc{(Hz)} &  $(\uparrow)$ & $(\uparrow)$ & $(\uparrow)$ & $(\uparrow)$ & $(\uparrow)$ & $(\uparrow)$ & $(\uparrow)$ & $(\uparrow)$ & $(\uparrow)$ & \textsc{($\uparrow$)} & \textsc{($\downarrow$)}	& \textsc{($\downarrow$)} & \textsc{($\downarrow$)} & \textsc{($\downarrow$)} \\
\midrule
\textsc{Random Agent} & 10 & 0.002 & 0.116 & 0.014 & 0.034 & 0.000 & 0.000 & 0.735 & 0.148 & 0.191 & 0.144 & 50.739 & 50.706 & 1.000 & 0.613 \\
\rowcolorize\textsc{Const. Velocity} & 10 & 0.043 & 0.067 & 0.252 & 0.439 & 0.202 & 0.355 & 0.739 & 0.455 & 0.451 & 0.381 & 7.923 & 7.923 & 0.314 & 0.293 \\
\textsc{MTR+++} \cite{Qian23tr_SimpleEffectiveSimMultiAgent}& 2 & 0.321 & 0.247 & 0.428 & 0.533 & 0.340 & 0.886 & 0.797 & 0.655 & 0.893 & 0.700 & \textbf{2.125} & 1.679 & 0.080 & \textbf{0.135} \\
\rowcolorize\textsc{MVTE} \cite{Wang23tr_MultiverseTransformer}& 10 & 0.353 & 0.354 & 0.496 & 0.599 & \textbf{0.392} & 0.913 & \textbf{0.833} & 0.642 & \textbf{0.907} &	0.732 &	3.859 &	1.674 &	0.090 &	0.158 \\
\textsc{Trajeglish} \cite{Philion24iclr_Trajeglish}& 10 & \textbf{0.356} & \textbf{0.399} & \textbf{0.509} & \textbf{0.654} & 0.378 & \textbf{0.925} & \textbf{0.834} & \textbf{0.660} & 0.884 &	\textbf{0.735} &	3.158 &	\textbf{1.615} &	0.076 &	0.170 \\
\midrule
\rowcolorize \textcolor{waymoblue}{\textsc{SceneDMF}} \cite{Guo23arxiv_SceneDM}& 0.125 & 0.343 & 0.395 & 0.381 & 0.366 & 0.362 & 0.760 & 0.812 & 0.623 & 0.735 & 0.628 & 4.158 & 2.414 & 0.217 & 0.285 \\
\textcolor{waymoblue}{\textsc{VBD}} \cite{Huang24arxiv_VBD}& 0.125 & 0.359 & 0.366 & 0.420 & 0.522 & 0.368 & 0.934 & 0.815 & 0.651 & 0.879 & 0.720 & 2.257 & 1.474 & 0.036 & 0.152 \\
\rowcolorize \textcolor{waymoblue}{\textsc{Ours (One-Shot)}}& 0.125 & 0.399 & 0.225 & 0.510 & 0.605 & 0.401 & 0.934 & 0.837 & 0.686 & 0.893 & 0.736 & 2.436 & 1.103 & 0.070 & 0.172 \\
\midrule
\textcolor{waymoblue}{\textsc{Ours (Amortized)}}& 10 & 0.310 & 0.389 & 0.459 & 0.560 & 0.349 & \textbf{0.917} & 0.815 & 0.634 & 0.833 & 0.703 & 2.619 & 1.767 & \textbf{0.064} & 0.177 \\
\midrule
 \bluerowcolorize\textsc{Logged Oracle}& - & 0.476 & 0.478 & 0.578 & 0.694 & 0.476 & 1.000 & 0.883 & 0.715 & 1.000 & 0.819 & 0.000 & 0.000 & 0.028 & 0.111 \\
    \bottomrule
    \end{tabular}
    \endgroup
    \end{adjustbox}
    \label{tab:wosac-per-metric-scores-test}
\end{table*}

\vspace{-.5em}
\subsection{Scene Generation}
\vspace{-.5em}

\boldparagraph{Unconstrained Scene Generation} We use the unconditional scene generation task as a means to quantitatively measure the distributional realism of our model. We condition the scene using the same logged road graph and traffic signals, as well as the logged agent validity to control for the same number of agents generated per scene. All agent attributes are generated by the model.

Due to a lack of public benchmarks for this task, we adopt a slightly modified version of the WOSAC \cite{Montali23neurips_wosac} metrics, where different metrics buckets are aggregated per-scene instead of per-agent, due to the lack of one-to-one correspondence between agents in the generated scene versus the logged scene (see Appendix \ref{sec:scenegen-metrics-details} for more details). Metrics are aggregated over all agents that are ever valid in the 9 second trajectory.

We show our model's realism metrics in Tab.~\ref{tab:scenegen_womd}. Even compared to the oracle performance (comparing logged versus logged distributions), our model achieves comparable realism scores in every realism bucket. Introducing hard constraints on collisions can significantly improve the composite metric by preventing collisions, while scaling the model without hard constraints improves most realism metrics as the model learns to generate more realistic trajectories. The realism metrics only apply to trajectories and do not account for generated agent type and size distributions. We compare the generated size distributions versus log distributions in Fig.~\ref{fig:scenegen_distribution_comparison} and find the marginal and joint distributions both closely track the logged distribution. We show more examples of diverse, unconstrained scene generation when conditioning on the same global context in Appendix~\ref{subsec:scenegen} Fig.~\ref{fig:scenegen-grid}.

\boldparagraph{Constrained Scene Generation and Augmentation} The controllability we possess in the scene generation process as a product of our diffusion model design can be useful for targeted generation and augmentation of scenes. In Fig.~\ref{fig:exp_scenegen_viz}, we show qualitative results of scenes with constrained agents generated either via manually defined configs or by a few-shot prompted LLM. Extended qualitative results are listed in Appendix~\ref{subsec:appendix_scenegen_results}.

\subsection{Model Design Analysis and Ablation Studies}
\label{sec:scaling}

\begin{figure}[t!]%
\vspace{-4mm}
\begin{floatrow}
\ffigbox[.42\textwidth]
{%
    \vspace{-2mm}
    \centering
    \input{tikz/constrained_scenegen}
    \vspace{-4mm}
}
{%
  \caption{Long-tail synthetic scenes generated via control points either explicitly defined by a manually defined (M) or LLM-generated (L) config. \textcolor{magenta}{Magenta} indicates generated motorcyclists/car agents.}
    \label{fig:exp_scenegen_viz}
}
\ffigbox[.56\textwidth]
{%
    \centering
    \newcommand{\trimAmount}{1em} %

\adjustbox{width=\linewidth, valign=m}{
\setlength{\tabcolsep}{0pt} %
\begin{tabular}{@{}ccccc@{}} %
  & 8 & 4 & 2 & 1 \\ %
S & \raisebox{-.5\height}{\includegraphics[width=0.25\linewidth, trim={\trimAmount} {\trimAmount} {\trimAmount} {\trimAmount},clip]{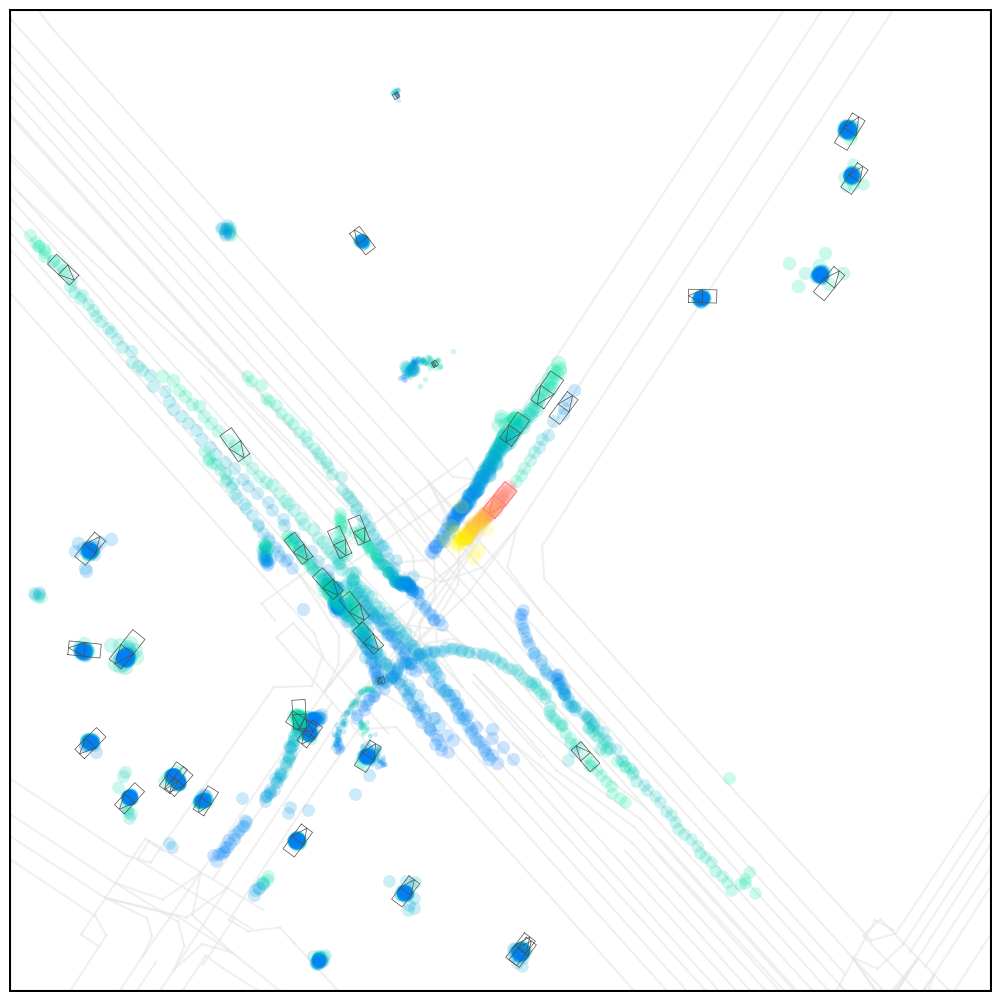}} &
  \raisebox{-.5\height}{\includegraphics[width=0.25\linewidth, trim={\trimAmount} {\trimAmount} {\trimAmount} {\trimAmount},clip]{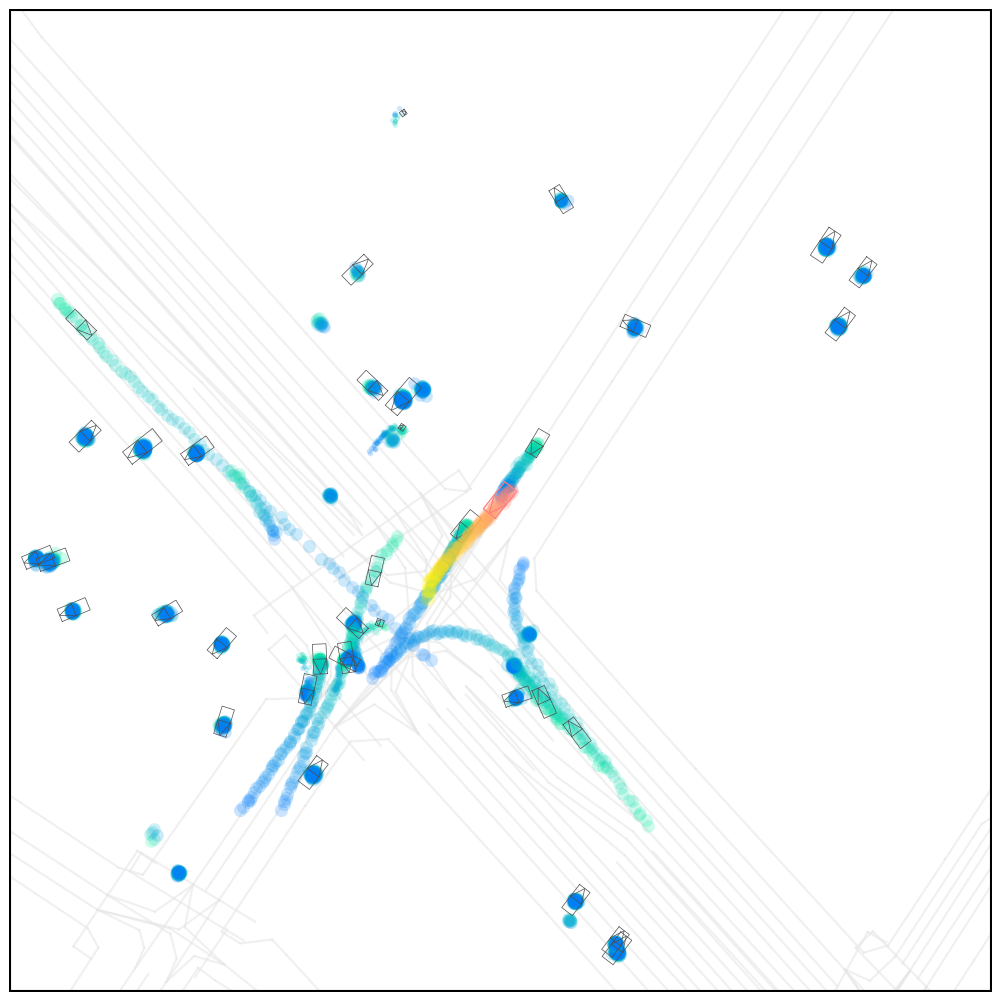}} &
  \raisebox{-.5\height}{\includegraphics[width=0.25\linewidth, trim={\trimAmount} {\trimAmount} {\trimAmount} {\trimAmount},clip]{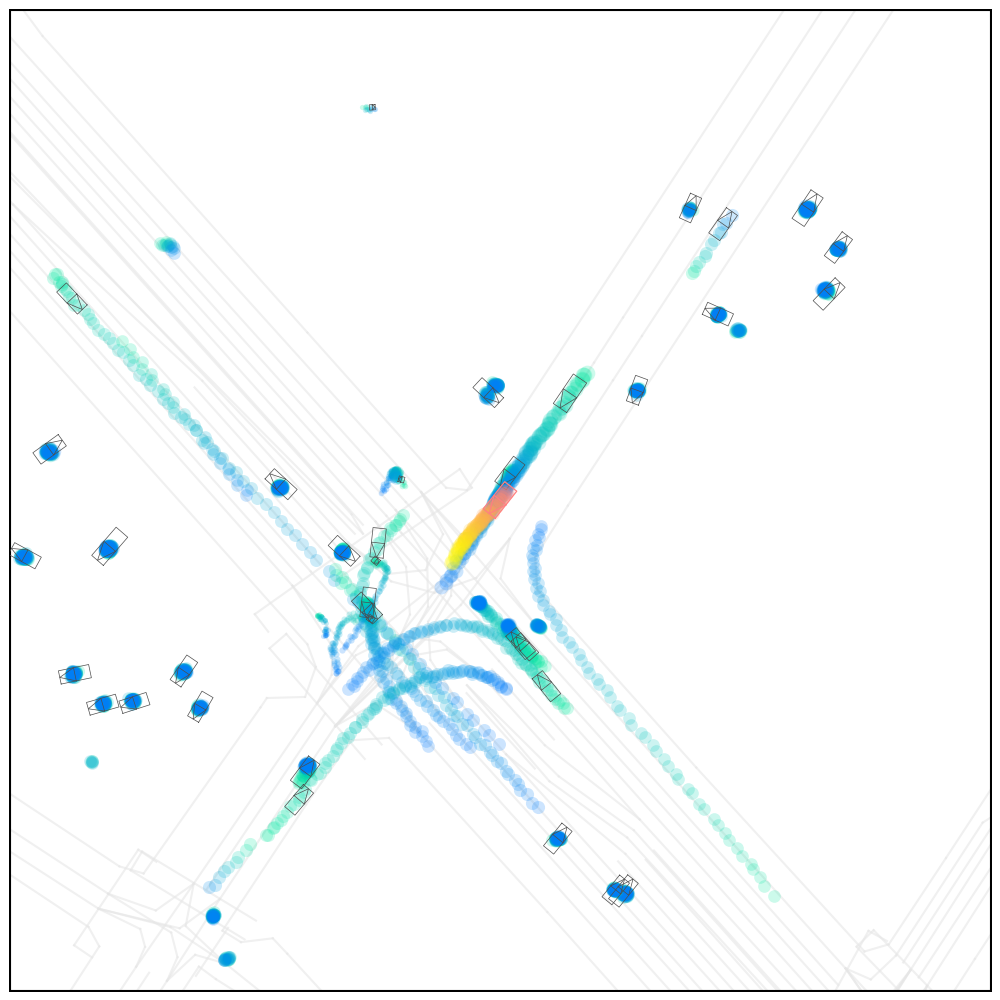}} &
  \raisebox{-.5\height}{\includegraphics[width=0.25\linewidth, trim={\trimAmount} {\trimAmount} {\trimAmount} {\trimAmount},clip]{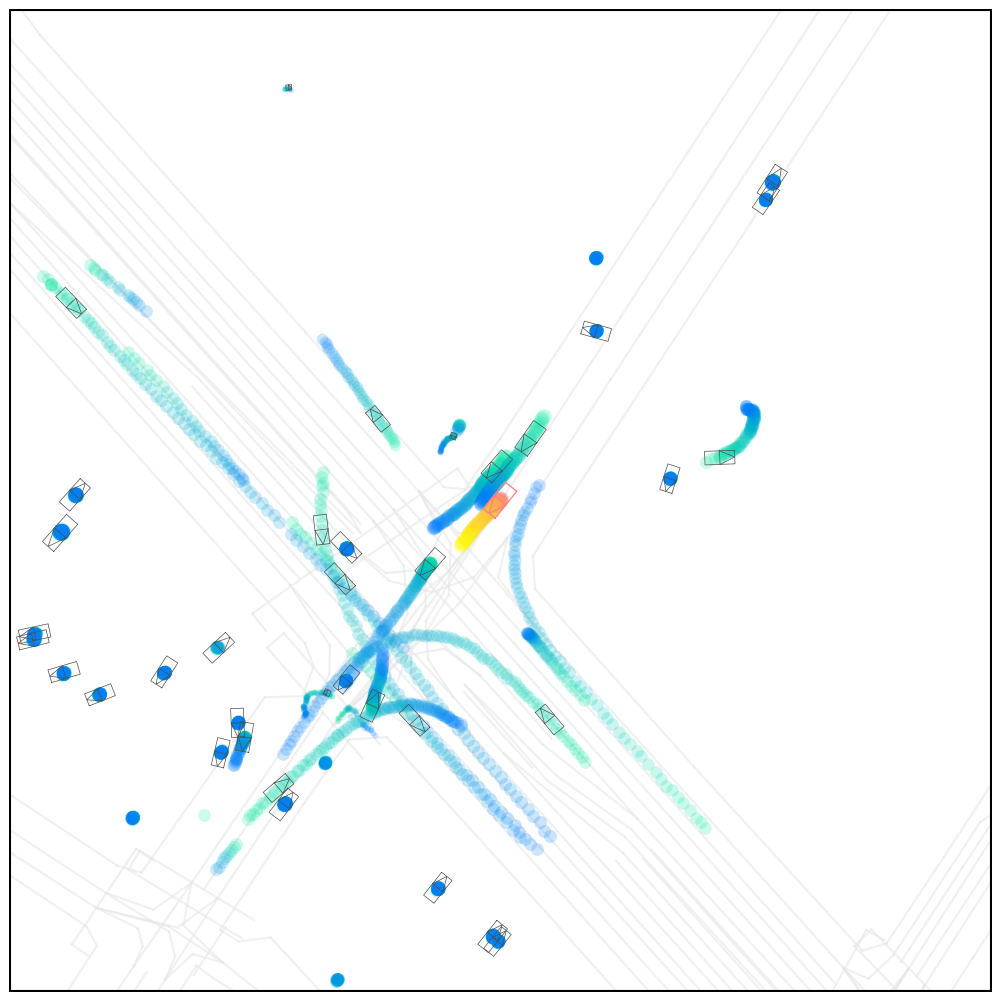}} \\
M & \raisebox{-.5\height}{\includegraphics[width=0.25\linewidth, trim={\trimAmount} {\trimAmount} {\trimAmount} {\trimAmount},clip]{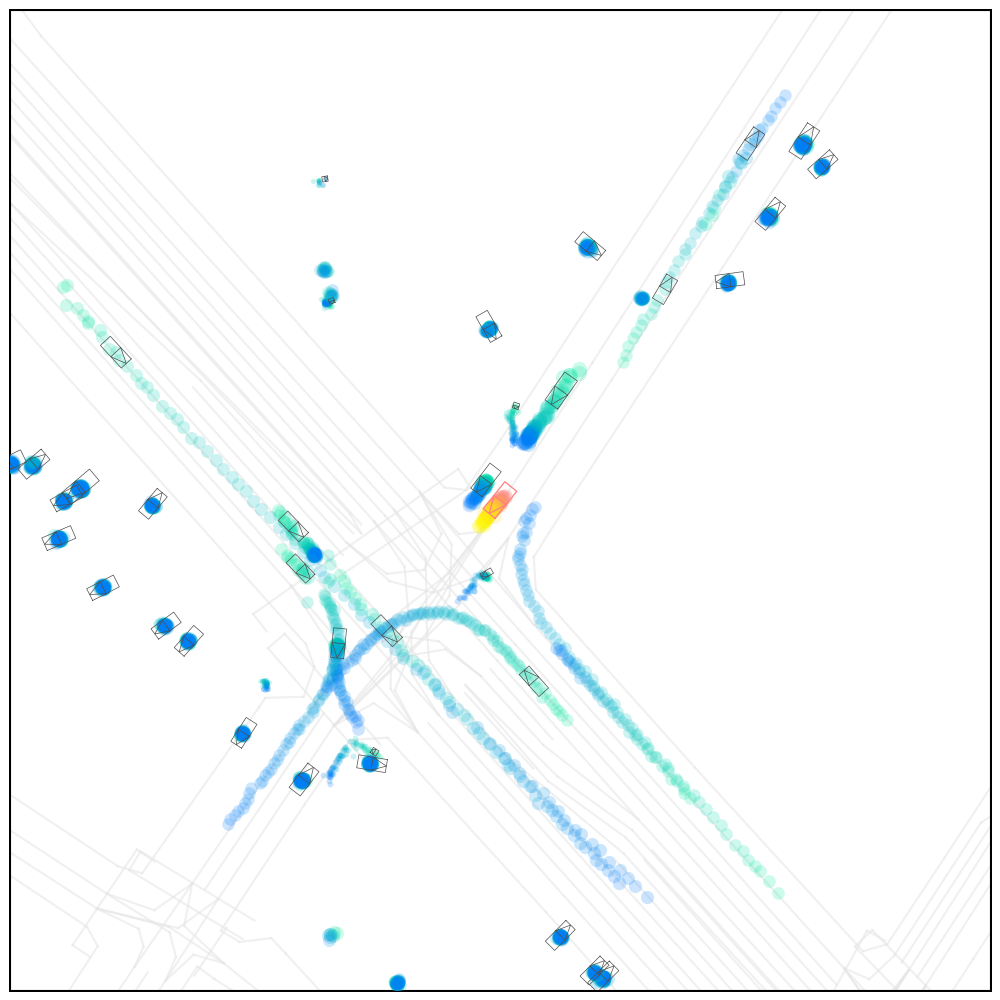}} &
  \raisebox{-.5\height}{\includegraphics[width=0.25\linewidth, trim={\trimAmount} {\trimAmount} {\trimAmount} {\trimAmount},clip]{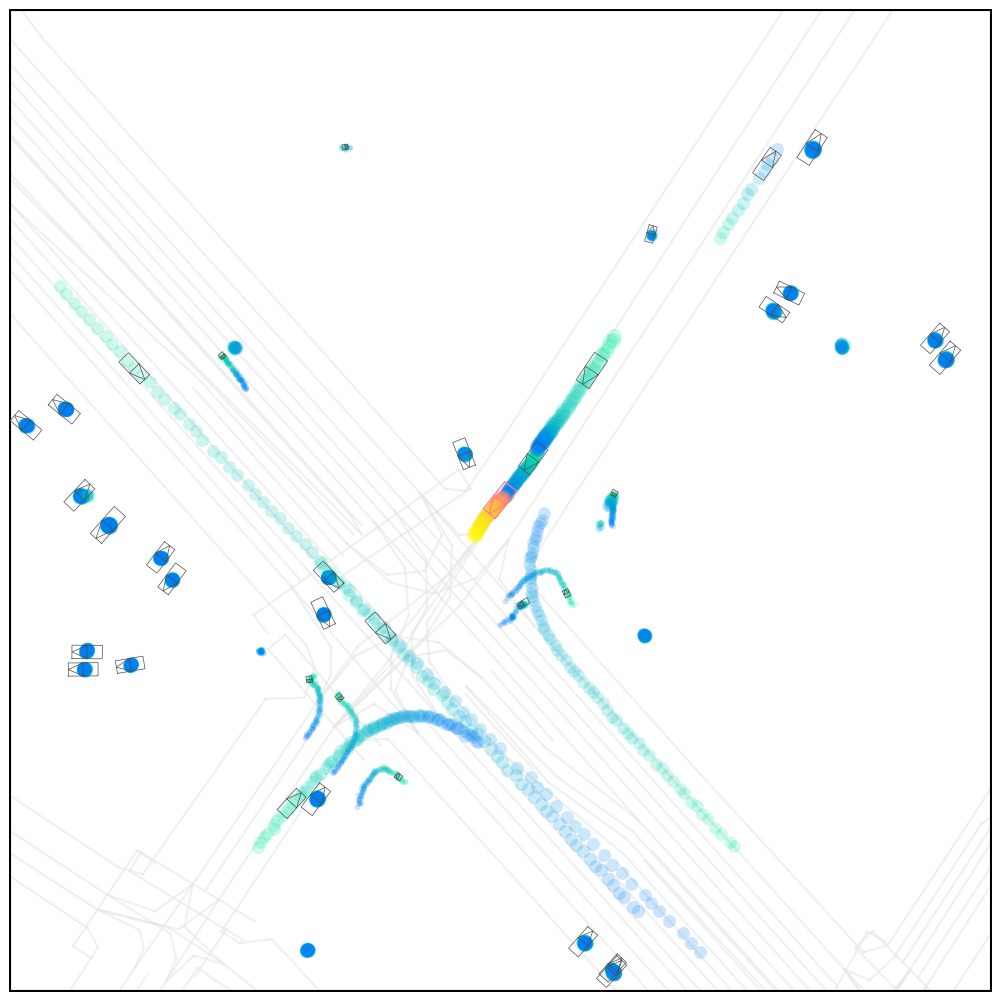}} &
  \raisebox{-.5\height}{\includegraphics[width=0.25\linewidth, trim={\trimAmount} {\trimAmount} {\trimAmount} {\trimAmount},clip]{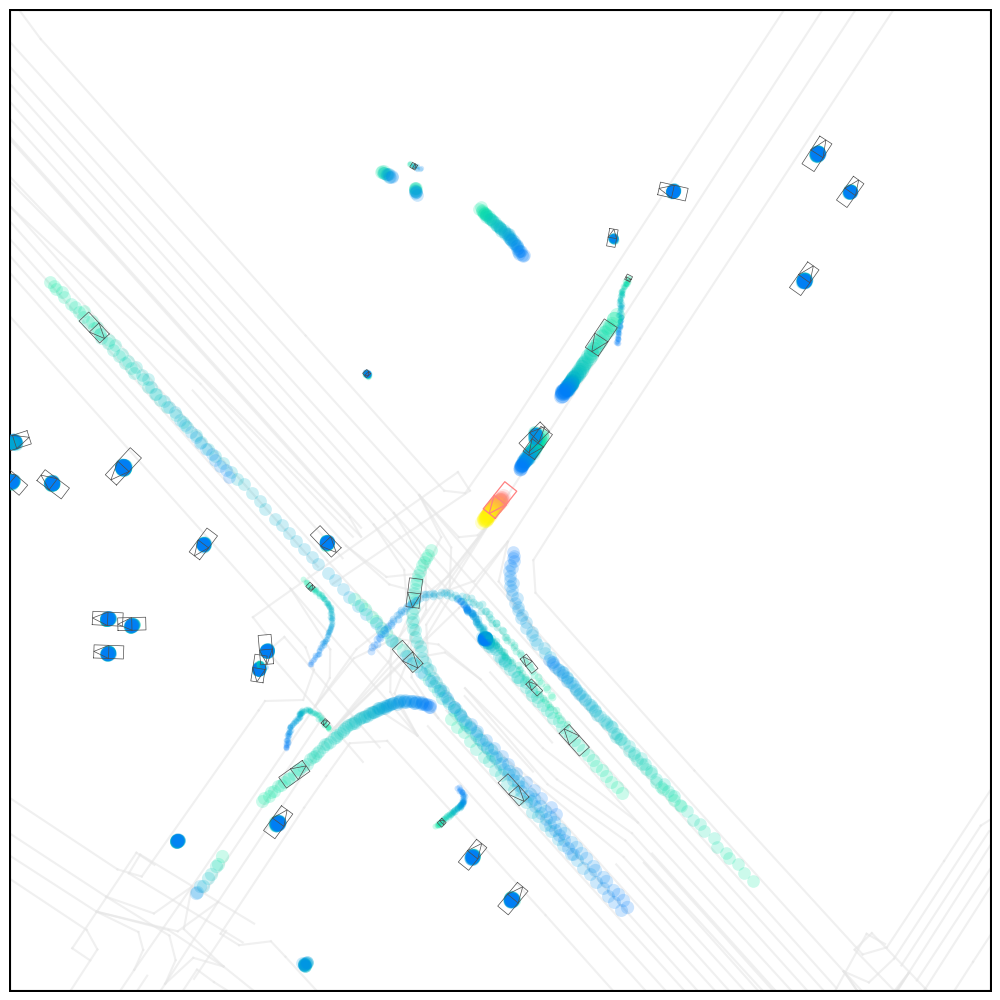}} &
  \raisebox{-.5\height}{\includegraphics[width=0.25\linewidth, trim={\trimAmount} {\trimAmount} {\trimAmount} {\trimAmount},clip]{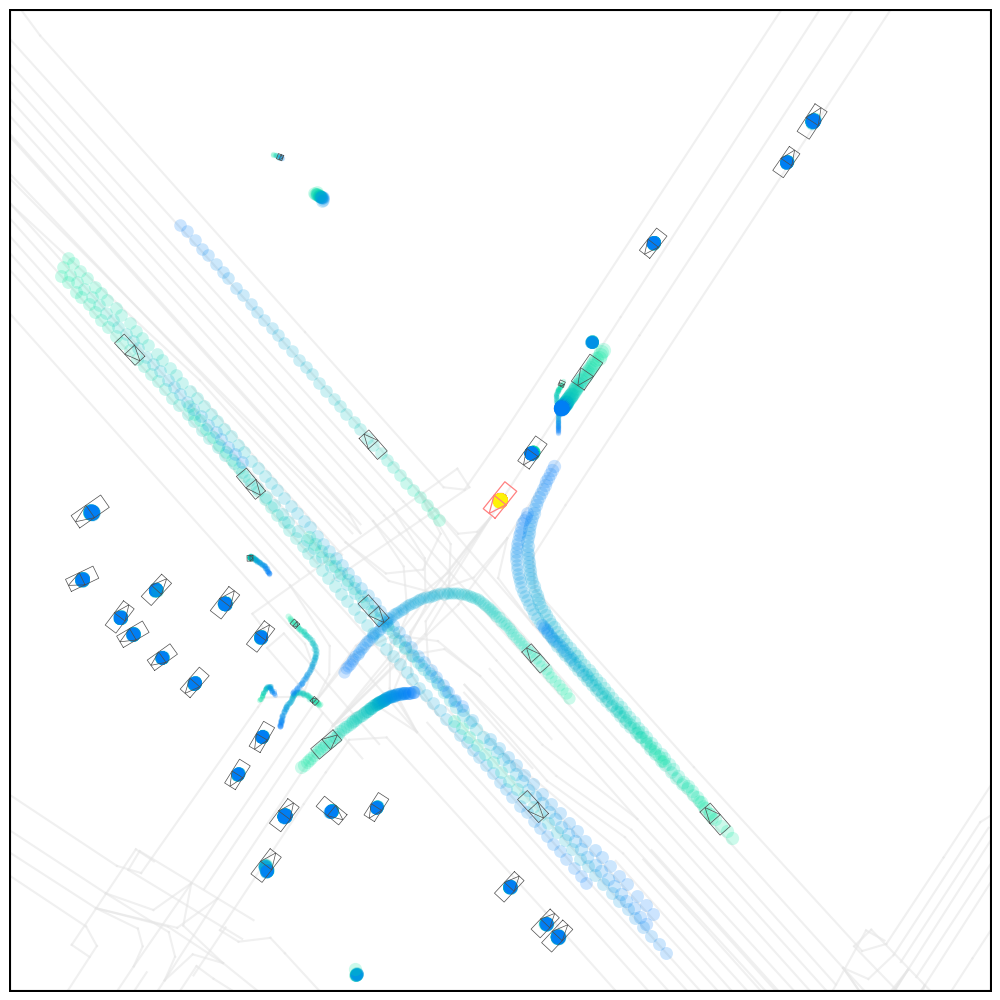}} \\
L & \raisebox{-.5\height}{\includegraphics[width=0.25\linewidth, trim={\trimAmount} {\trimAmount} {\trimAmount} {\trimAmount},clip]{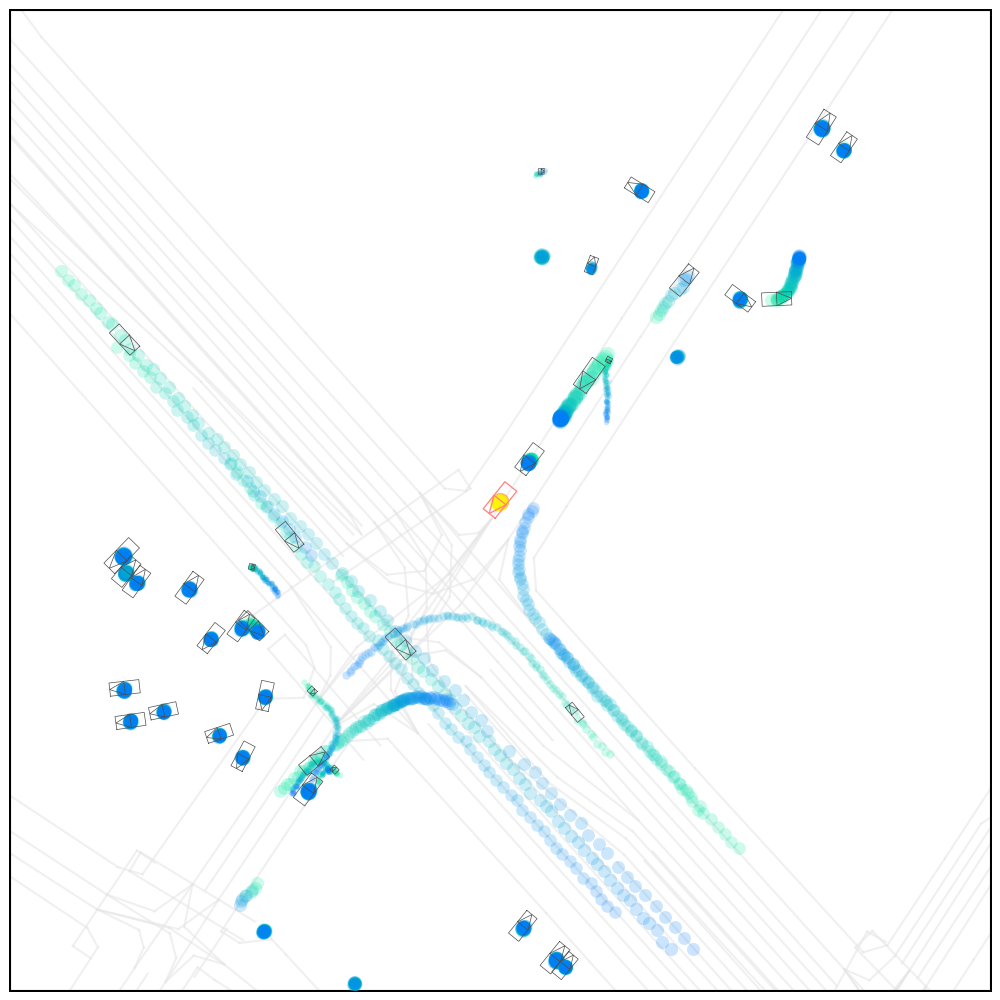}} &
  \raisebox{-.5\height}{\includegraphics[width=0.25\linewidth, trim={\trimAmount} {\trimAmount} {\trimAmount} {\trimAmount},clip]{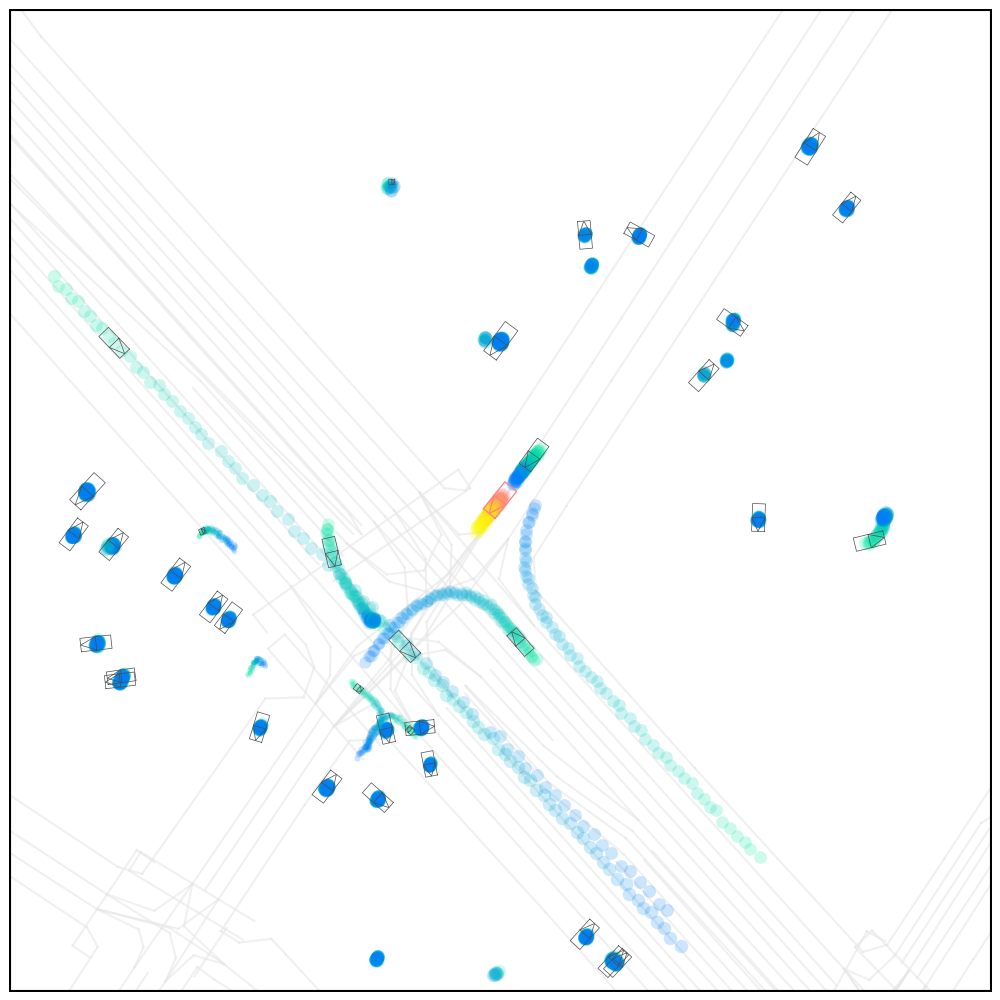}} &
  \raisebox{-.5\height}{\includegraphics[width=0.25\linewidth, trim={\trimAmount} {\trimAmount} {\trimAmount} {\trimAmount},clip]{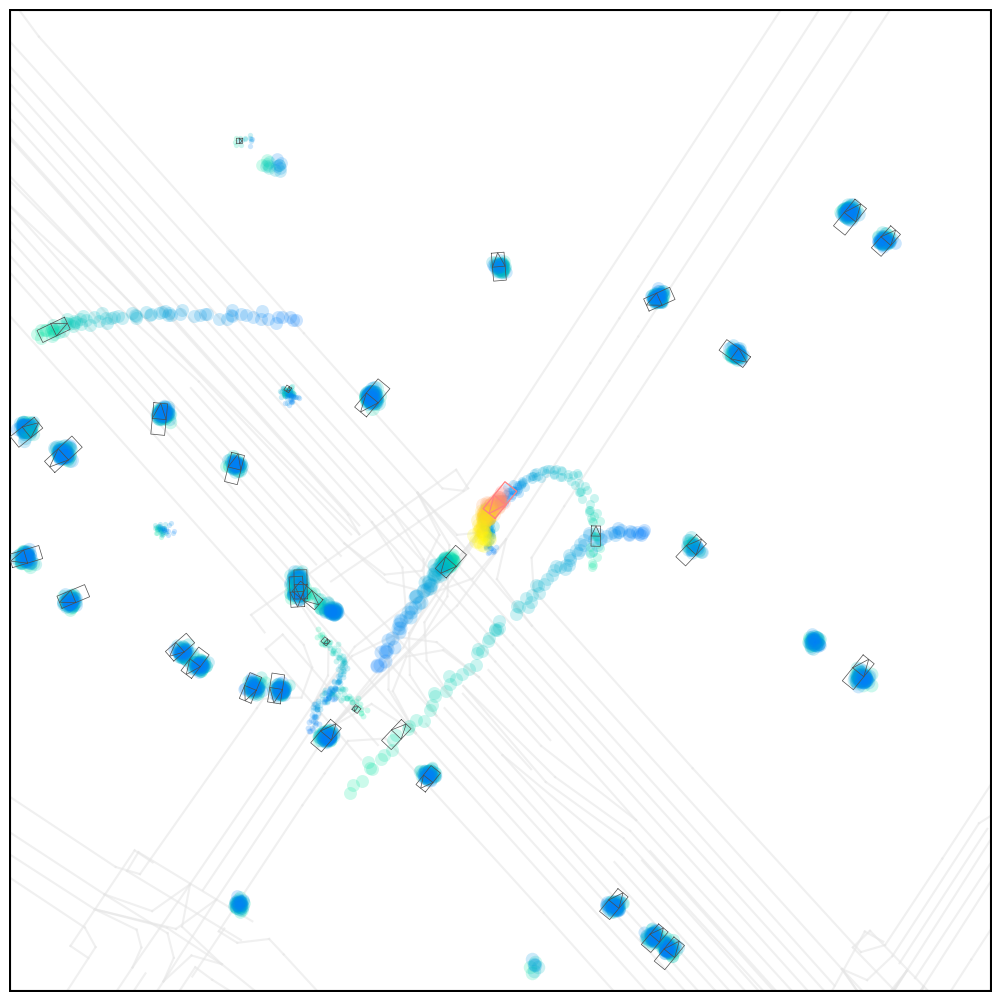}} & %
  \raisebox{-.5\height}{\includegraphics[width=0.25\linewidth, trim={\trimAmount} {\trimAmount} {\trimAmount} {\trimAmount},clip]{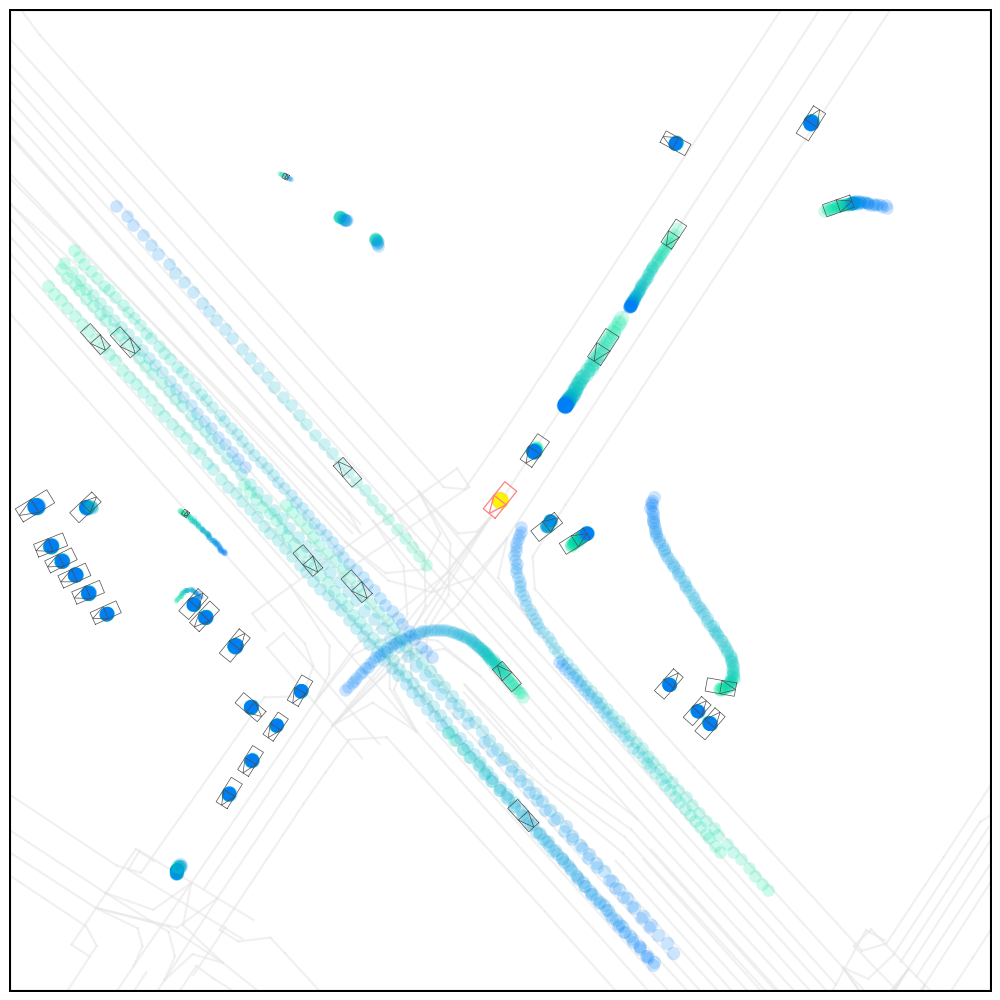}} \\
\end{tabular}
}

}
{%
    \caption{Fully synthetic scene generation quality comparison via scaling model parameters ($S\rightarrow L$) and increasing temporal resolution (patch size $8\rightarrow 1$). Increasing compute by scaling either the model size or temporal resolution improves the overall realism.}
    \label{fig:visualize-model-scaling}
}
\end{floatrow}
\end{figure}

\textbf{Scaling Analysis}
Given two options of scaling model compute, either by increasing transformer temporal resolution by decreasing temporal patch sizes, or increasing the number of model parameters, we investigate the performance of multiple transformer backbones: \{Model Size\} $\times$ \{Temporal Patch Size\} = \{L, M, S\} $\times$ \{8, 4, 2, 1\}. We vary model size by jointly scaling the number of transformer layers, hidden dimensions, and attention heads (see Sec.~\ref{sec:scaling-hyperparams} of Appendix for details). We show quantitative results from this model scaling in Fig.~\ref{fig:model-scaling} and qualitative comparisons in Fig.~\ref{fig:visualize-model-scaling}. Increasing both temporal resolution and number of model parameters improves realism of the simulation.

\boldparagraph{Multi-task Compatibility} We find that multitask co-training across BP, SceneGen and with random control masks improves performance compared to a single-task, BP only model on the sim agent rollout task, notably reducing collision and offroad rates. We find that jointly learning multiple agent features ($x, y, z, \gamma$, size, type) achieves on-par performance with a pose-only ($x, y, z, \gamma$) model.

\boldparagraph{Model Architecture Ablation} As shown in Tab.~\ref{tab:ablation}, replacing AdaLN-Zero conditioning with cross attention leads to a 7.99\% decrease in realism performance, largely due to significantly higher collision rates and offroad rates. Removing the agent-wise spatial attention layer very significantly increases collision rate, as it removes the mechanism for agents to learn a joint distribution.

\section{Conclusion}

We have introduced \ours{}, a scene-level diffusion prior designed for traffic simulation. \ours{} combines scene initialization with scene rollout to provide a diffusion-based approach to closed-loop agent simulation that is efficient (through amortized autoregression) and controllable (through generalized hard constraints). We performed scaling and ablation studies and demonstrated model improvements with computational resources. On WOSAC, we demonstrate competitive results with the leaderboard and state-of-the-art performance among diffusion methods.

\boldparagraph{Limitations} While our amortized diffusion approach is, to our knowledge, the only and best performing closed-loop diffusion-based agent model with competitive performance, we do not exceed current SOTA performance for other autoregressive models. We do not explicitly model validity masks and resort to logged validity in this work. Future work looks to also model the validity mask.

\boldparagraph{Broader Impact} This paper aims to improve AV technologies. With our work we aim to make AVs safer by providing more realistic and controllable simulations. The generative scene modeling techniques developed in this work could have broader social implications regarding generative media and content generation, which poses known social benefits as well as risks of misinformation.

\begin{ack}
No third-party funding received in direct support of this work. We thank Shimon Whiteson for his detailed for detailed feedback and also the anonymous reviewers. We would like to thank Reza Mahjourian, Rongbing Mu, Paul Mougin, and Nico Montali for offering consultation and feedback on evaluation metrics. We thank Kevin Murphy for his assistance in developing the mathematical notation for the likelihood metrics. We thank Zhaoyi Wei and Carlton Downey for helpful discussions about the project. All the authors are employees of Waymo LLC.
\end{ack}

\clearpage{}

{
\small
\bibliographystyle{plainnat}
\bibliography{main}

}

\clearpage{}
\appendix

\section{Appendix / supplemental material}

\subsection{WOSAC Metrics}
\label{sec:wosac-metrics-details}

Suppose there are $N\approx 500k$ scenarios, each of length $T=80$ steps, each containing $A \leq 128$ agents (objects). For each scenario, we generate $K=32$ samples (conditioned on the true initial state), which is a set of trajectories for each object for each time step, where each point in the trajectory is a $D=4$-dim vector recording  location $(x,y,z)$ and orientation $\theta$. Let all this generated data be denoted by $x$($1:N$, $1:A$, $1:K$, $1:T$, $1:D$). Let the ground truth data be denoted $x^*$($1:N$, $1:A'$, $1:T$, $1:D$). Below we discuss how to evaluate the likelihood of the true (test) dataset $x^*$ under the  distribution induced by the simulated dataset $x$.

(Note that we may have $A' > A$, since the ground truth (GT) can contain cars that enter the scene after the initial prefix used by the simulator; this is handled by defining a validity mask, $v(1:N, 1:T, 1:A')$, which is set to 0 if we want to exclude a GT car from the evaluation, and is set to 1 otherwise.)

Rather than  evaluating the realism of the full trajectories in the raw $(x,y,z,\theta)$ state space, WOSAC defines $M=9$ statistics (scalar quantities of interest) from each trajectory. Let $F_j(x(i,a,:))$ represent the set of statistics/features (of type j) derived from  $x(i, a, 1:K, 1:T)$ by pooling over $T,K$. This is used to compute a histogram $p_{ija}(.)$ for the empirical distribution of $F_j$ for scenario i. Let $F_j(x^*(i,a,t))$ be the value of this statistic from the true trajectory $i$ for vehicle $a$ at time $t$ . Then we define the negative log likelihood to be \begin{equation}
 NLL(i,a,t,j) = -\log p_{ija}(F_j(x^*(i,a,t))
\end{equation}
The j'th metric for scenario i is defined as 
\begin{equation}
\begin{aligned}
  m(a,i,j) &= \exp\Big(- [\frac{1}{N(i,a)}] \sum_t v(i,a,t) NLL(i,a,t,j) \Big) \\
   m(i,j) &= \frac{1}{A} \sum_a m(a,i,j) \\
   N(i,a) &= \sum_t v(i,a,t) \text{ is the number of valid points}.
\end{aligned}
\end{equation}
Finally an aggregated metric, used to rank entries, is computed as 
\begin{equation}
  score =\frac{1}{N'} \frac{1}{M} \sum_{i=1}^{N'} \sum_{j=1}^M w_j m(i,j)
\end{equation}
where $0 \leq w_j \leq 1$.

The 9 component metrics are defined as linear speed, linear acceleration magnitude, angular speed, angular acceleration magnitude, distance to nearest object, collisions, time-to-collision (TTC), distance to road edge, and road departures.

\subsection{SceneGen Metrics}
\label{sec:scenegen-metrics-details}

We instead let $F_j(x(i,:))$ represent the set of statistics/features (of type j) derived from  $x(i, 1:A', 1:K, -H:T)$ by pooling over $T,A',K$. This is used to compute a histogram $p_{ij}(.)$ for the empirical distribution of $F_j$ for scenario i.

\subsection{Additional Evaluation Details}
\boldparagraph{Simulation step validity} Due to the requirement of validity masks during inference, which is applied as an attention padding mask within the transformer, the model does not generate valid values for invalid steps. As the WOSAC challenge evaluates simulation agents for all steps, regardless of the step's validity, we use linear interpolation / extrapolation to impute values for all invalid steps in our simulations for the final evaluation.

\subsection{Additional Dataset Information}

WOSAC uses the v1.2.0 release of WOMD, and we treat WOMD as a set $\mathcal{D}$ of scenarios where each scenario is a history-future pair. This dataset offers a large quantity of high-fidelity object behaviors and shapes produced by a state-of-the-art offboard perception system. We use WOMD’s 9 second 10 Hz sequences (comprising H = 11 observations from 1.1 seconds of history and 80 observations from 8 seconds of future data), which contain object tracks at 10 Hz and map data for the area covered by the sequence. Across the dataset splits, there exists 486,995 scenarios in train, 44,097 in validation, and 44,920 in test.

\subsection{Additional Amortized Diffusion Algorithm Details}

\boldparagraph{Warm up: } At inference time, the rollout process is preceded by a warm up step. The warm up step is necessary for initializing a buffer of future timesteps before any diffusion iterations take place. The warm up entails a single iteration of a one-shot prediction process described in Algorithm~\ref{algo:one-shot}. This process samples pure noise for some future steps and conditions the denoising process on the set of past steps.

\boldparagraph{Amortized autoregressive rollout: }
In Fig.~\ref{fig:schematic_amortized}, we provide a visual illustration of our amortized autoregressive rollout procedure. We operate the rollout procedure using a buffer to track future steps in the trajectory. After the warm up, the future buffer contains $\Tau$ predicted steps with an increasing noise level. Note that step $\tau=1$ has very little noise applied. The future buffer in this state is denoised for a single iteration using past steps to condition the process. After a single iteration, the clean step at $\tau=1$ is popped off of the buffer, and it is added to the past steps. Before the next iteration, a step $\tau=\Tau+1$ is sampled from a pure noise distribution and is appended to the end of the future buffer. The described rollout process can be repeated to generate trajectories of arbitrary length as clean steps are popped off the buffer. 

\subsection{Additional Implementation Details}

\boldparagraph{Architecture Details}: For our base model, our scene encoder architecture uses 256 latent queries.  Each scene token is 256-dimensional, with 4 transformer layers and 4 transformer heads, with a transformer model dimension of 256. We train and run inference with all 128 agents. 

\label{sec:scaling-hyperparams}
\boldparagraph{Scaling Hyperparameters:}\\
\emph{Small Model}: Scene token dimension 128, 2 Transformer layers, 128 Transformer model dimensions, 2 Transformer Heads.

\emph{Medium Model}: Scene token dimension 256, 4 Transformer layers, 256 Transformer model dimensions, 4 Transformer Heads.

\emph{Large Model}: Scene token dimension 512, 8 Transformer layers, 512 Transformer model dimensions, 8 Transformer Heads.

\boldparagraph{Optimizer}: We use the Adafactor optimizer \cite{Shazeer18icml_Adafactor}, with EMA (exponential moving average). We decay using Adam, with $\beta_1=0.9$, $\text{decay}_{adam}=0.9999$, weight decay of 0.01, and clip gradient norms to $1.0$. 

\boldparagraph{Training details} Train batch size of 1024, and train for 1.2M steps. We select the most competitive model based on validation set performance, for which we perform a final evaluation using the test set. We use an initial learning rate of $3 \times 10^{-4}$. We use 16 diffusion sampling steps. When training, we mix the behavior prediction (BP) task with the scene generation task, with probability 0.5. The randomized control mask is applied to both tasks.

\label{subsec:normalization}
\boldparagraph{Additional Hyperparameters} To preprocess features, we use scaling constants of $\frac{1}{80}$ for features $x,y,z$, and compute mean $\mu$ and standard deviation $\sigma$ of features $l, w, h$.

We preprocess each agent feature $f$ to produce normalized feature $f^\prime$ via $f^\prime = \frac{f - \mu_f}{2 * \sigma_f}$, where:
\begin{align}
    \mu_l = 4.5, \quad \mu_w = 2.0, \quad \mu_h = 1.75, \quad \mu_k = 0.5 .
\end{align}
and
\begin{align}
    \sigma_l = 2.5, \quad \sigma_w = 0.8, \quad \sigma_h = 0.6, \quad \sigma_k = 0.5.
\end{align}
We scale by twice the std $\sigma$ values to allow sufficient dynamic range for high feature values for some channels.

\begin{figure}
    \centering
    \input{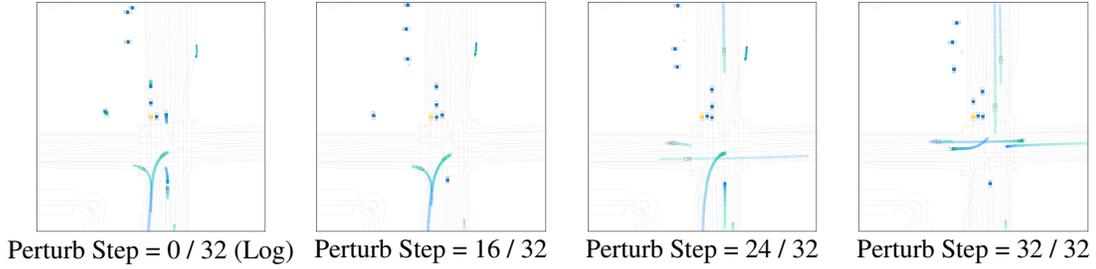}
    \caption{Log-perturbation via noising and denoising the original logs to different noise levels. An increasing level of noise added and then removed results in scenes more and more dissimilar from the original log, yet increasingly diverse. The scenes are realistic regardless of the perturbation noise level.}
    \label{fig:exp_log_perturbation}
\end{figure}

\subsection{Prompts used in Language-based few-shot Scene Generation}
\label{subsec:prompts}

\subsubsection{Prompt:}

\renewcommand\lstlistingname{Prompt}
\begin{lstlisting}[breaklines=true,caption={Prompts used in Language-based few-shot Scene Generation.}]
You are writing proto to generate and control an agent's behavior 
for an autonomous vehicle (AV) simulation. I will give the follow 2
examples of input and the generated proto MultiAgentMultiConstraint, 
which will constrain the agent's position in either past, current, or
future timestep. 5 timesteps equals 1 second, so for example 15 steps
would equal 3 seconds. Given a natural language description of the
agent's desired behavior, please generate the corresponding
MultiAgentMultiConstraint.

Very Important limitations:
1. Only time_step_idx values in [0,8] are valid for PAST time step.
2. Only time_step_idx values in [0] are valid for CURRENT time step.
3. Only time_step_idx values in [0,49] are valid for FUTURE time step.
4. You may only use types POT_CAR, POT_MOTORCYCLIST, POT_PEDESTRIAN to generate these examples
5. No two agents should overlap each other at the same time step in the same time frame.

The following are 2 examples of a natural language input and the
output is a text file that creates the corresponding constraint.

Example 1
Input: Generate constraints where a motorcycle agent cuts in front of
the AV coming from the right side at some time in the future.
Output:
```
agent_constraints {
  step_constraints {
    relative_step_constraint {
      time_frame: CURRENT
      time_step_idx: 0
      lat_distance: 3.7 # Slightly right of ADV
      long_distance: 0.0
      agent_type: POT_MOTORCYCLIST
    }
  }
  step_constraints {
    relative_step_constraint {
      time_frame: FUTURE
      time_step_idx: 30
      lat_distance: 0.0
      long_distance: 3.0 # ahead of AV
      agent_type: POT_MOTORCYCLIST
    }
  }
  agent_name: 'motorcycle_0'
}
scene_name: 'single_motorcycle_cut_in'
```

Example 2
Input: Generate constraints where a car agent is tailgating the AV by
following behind it closely.
Output:
```
agent_constraints {
  step_constraints {
    relative_step_constraint {
      time_frame: PAST
      time_step_idx: 0
      lat_distance: 0.0
      long_distance: -15.0
      agent_type: POT_CAR
    }
  }

  step_constraints {
    relative_step_constraint {
      time_frame: CURRENT
      time_step_idx: 0
      lat_distance: 0.0
      long_distance: -10.0
      agent_type: POT_CAR
    }
  }
  step_constraints {
    relative_step_constraint {
      time_frame: FUTURE
      time_step_idx: 30
      lat_distance: 0.0
      long_distance: -2.0
      agent_type: POT_CAR
    }
  }
  agent_name: 'car_0'
}
scene_name: 'single_car_tailgater'
```

Please output just the MultiAgentMultiConstraint in the response and
leave any explanation in the comments. Please double check the 
important limitations described before are met.

Now take a deep breath and lets think step by step to write the proto
given the following  input:

S-Shaped Maneuver: Create a motorcyclist agent that is doing a
S-shaped maneuver around the AV.

Surrounding Traffic: Create a scene with 8 agents that are surrounding 
the AV and traveling along for a few seconds. There are 4 agents on 
the left lane of AV in a straight line and 4 agents on the lane of AV
in a straight line. 2 of the front agents at motorcyclists while the
rest are cars.
\end{lstlisting}

\subsubsection{Scene Diffusion Constraint Protos Result:}
\label{subsec:appendix_constraints}

\begin{lstlisting}[breaklines=true,caption={(Manual) Cut-in Constraint}]
agent_constraints {
    step_constraints {
      relative_step_constraint {
        time_frame: PAST
        time_step_idx: 0
        lat_distance: 3.0
        long_distance: 0.0
        agent_type: POT_CYCLIST
      }
    }
    step_constraints {
      relative_step_constraint {
        time_frame: FUTURE
        time_step_idx: 9
        lat_distance: 0.0
        long_distance: 0.0
        agent_type: POT_CYCLIST
      }
    }
    step_constraints {
      relative_step_constraint {
        time_frame: FUTURE
        time_step_idx: 40
        lat_distance: -3
        long_distance: 10
        agent_type: POT_CYCLIST
      }
    }
    agent_name: 'cut_in_0'
}
\end{lstlisting}

\begin{lstlisting}[breaklines=true,caption={(Manual) Tailgate Constraint}]
agent_constraints {
    step_constraints {
      relative_step_constraint {
        time_frame: PAST
        time_step_idx: 0
        lat_distance: 0
        long_distance: -20
        agent_type: POT_CYCLIST
      }
    }
    step_constraints {
      relative_step_constraint {
        time_frame: FUTURE
        time_step_idx: 20
        lat_distance: 0.0
        long_distance: -10
        agent_type: POT_CYCLIST
      }
    }
    step_constraints {
      relative_step_constraint {
        time_frame: FUTURE
        time_step_idx: 40
        lat_distance: 0
        long_distance: -3
        agent_type: POT_CYCLIST
      }
    }
    agent_name: 'tailgater_agent_0'
}
\end{lstlisting}

\begin{lstlisting}[breaklines=true,caption={(LLM Generated) S-shaped Constraint}]
agent_constraints {
    step_constraints {
      # Initial position slightly to the right of the AV (time_step_idx 0)
      relative_step_constraint {
        time_frame: CURRENT
        time_step_idx: 0
        lat_distance: 2.0  # Adjust this value as needed for initial lateral offset
        long_distance: 0.0
        agent_type: POT_CYCLIST
      }
    }
    step_constraints {
      # Move the motorcycle diagonally to the left in the future (around 15 steps)
      relative_step_constraint {
        time_frame: FUTURE
        time_step_idx: 15
        lat_distance: -3.0  # Adjust this value for desired leftward movement
        long_distance: 5.0   # Adjust this value for desired forward movement
        agent_type: POT_CYCLIST
      }
    }
    step_constraints {
      # Move the motorcycle diagonally back to the right in the future (around 30 steps)
      relative_step_constraint {
        time_frame: FUTURE
        time_step_idx: 30
        lat_distance: 2.0  # Adjust this value for desired rightward movement
        long_distance: 10.0  # Adjust this value for desired forward movement
        agent_type: POT_CYCLIST
      }
    }
    agent_name: 'motorcycle_0'
}
\end{lstlisting}

\begin{lstlisting}[breaklines=true,caption={(LLM Generated) Surrounding Traffic Constraint}]
agent_constraints {
    step_constraints {
      relative_step_constraint {
        time_frame: CURRENT
        time_step_idx: 0
        lat_distance: -3.7
        long_distance: 10.0
        agent_type: POT_MOTORCYCLIST
      }
    }
    step_constraints {
      relative_step_constraint {
        time_frame: FUTURE
        time_step_idx: 25
        lat_distance: -3.7
        long_distance: 30.0
        agent_type: POT_MOTORCYCLIST
      }
    }
    agent_name: 'motorcycle_left_0'
}
agent_constraints {
    step_constraints {
      relative_step_constraint {
        time_frame: CURRENT
        time_step_idx: 0
        lat_distance: -3.7
        long_distance: 5.0
        agent_type: POT_CAR
      }
    }
    step_constraints {
      relative_step_constraint {
        time_frame: FUTURE
        time_step_idx: 25
        lat_distance: -3.7
        long_distance: 25.0
        agent_type: POT_CAR
      }
    }
    agent_name: 'car_left_0'
}
agent_constraints {
    step_constraints {
      relative_step_constraint {
        time_frame: CURRENT
        time_step_idx: 0
        lat_distance: -3.7
        long_distance: -5.0
        agent_type: POT_CAR
      }
    }
    step_constraints {
      relative_step_constraint {
        time_frame: FUTURE
        time_step_idx: 25
        lat_distance: -3.7
        long_distance: -1.0
        agent_type: POT_CAR
      }
    }
    agent_name: 'car_left_1'
}
agent_constraints {
    step_constraints {
      relative_step_constraint {
        time_frame: CURRENT
        time_step_idx: 0
        lat_distance: -3.7
        long_distance: -10.0
        agent_type: POT_CAR
      }
    }
    step_constraints {
      relative_step_constraint {
        time_frame: FUTURE
        time_step_idx: 25
        lat_distance: -3.7
        long_distance: -6.0
        agent_type: POT_CAR
      }
    }
    agent_name: 'car_left_2'
}
agent_constraints {
    step_constraints {
      relative_step_constraint {
        time_frame: CURRENT
        time_step_idx: 0
        lat_distance: 3.7
        long_distance: 10.0
        agent_type: POT_MOTORCYCLIST
      }
    }
    step_constraints {
      relative_step_constraint {
        time_frame: FUTURE
        time_step_idx: 25
        lat_distance: 3.7
        long_distance: 30.0
        agent_type: POT_MOTORCYCLIST
      }
    }
    agent_name: 'motorcycle_right_0'
}
agent_constraints {
    step_constraints {
      relative_step_constraint {
        time_frame: CURRENT
        time_step_idx: 0
        lat_distance: 3.7
        long_distance: 5.0
        agent_type: POT_CAR
      }
    }
    step_constraints {
      relative_step_constraint {
        time_frame: FUTURE
        time_step_idx: 25
        lat_distance: 3.7
        long_distance: 25.0
        agent_type: POT_CAR
      }
    }
    agent_name: 'car_right_0'
}
agent_constraints {
    step_constraints {
      relative_step_constraint {
        time_frame: CURRENT
        time_step_idx: 0
        lat_distance: 3.7
        long_distance: -5.0
        agent_type: POT_CAR
      }
    }
    step_constraints {
      relative_step_constraint {
        time_frame: FUTURE
        time_step_idx: 25
        lat_distance: 3.7
        long_distance: -1.0
        agent_type: POT_CAR
      }
    }
    agent_name: 'car_right_1'
}
agent_constraints {
    step_constraints {
      relative_step_constraint {
        time_frame: CURRENT
        time_step_idx: 0
        lat_distance: 3.7
        long_distance: -10.0
        agent_type: POT_CAR
      }
    }
    step_constraints {
      relative_step_constraint {
        time_frame: FUTURE
        time_step_idx: 25
        lat_distance: 3.7
        long_distance: -6.0
        agent_type: POT_CAR
      }
    }
    agent_name: 'car_right_2'
}
\end{lstlisting}

\subsubsection{Controllable Scenegen Results}
\label{subsec:appendix_scenegen_results}

Qualitative results showing one successful and one failed example of applying the control points to scene generation task with the protos are listed in Appendix~\ref{subsec:appendix_constraints}. For measuring the success and failures of this scene generation task, we randomly selected 25 examples that were generated with each of the 4 control protos and qualitatively determined success on 1) if the new object does not overlap with any existing objects in the scene and 2) if the new object semantically behaves in the way intended by the control points. Otherwise, we considered it a failure. Overall, we measured a success rate of 40/100.

\begin{tikzpicture}[
    node distance=0.5cm,
    every node/.style={anchor=west},
    box/.style = {draw=#1, line width=0.5mm,inner sep=0.25mm}
]
\def\imagewidth{200pt} %
\def\rowsep{30pt}

\node (cut-in) at (0,0) {\textbf{Cut-in  (left: success, right: failure)}};

\node[below=4pt of cut-in.south west, inner sep=0pt, anchor=north west] (cut-in-good) {\includegraphics[width=\imagewidth,height=\imagewidth]{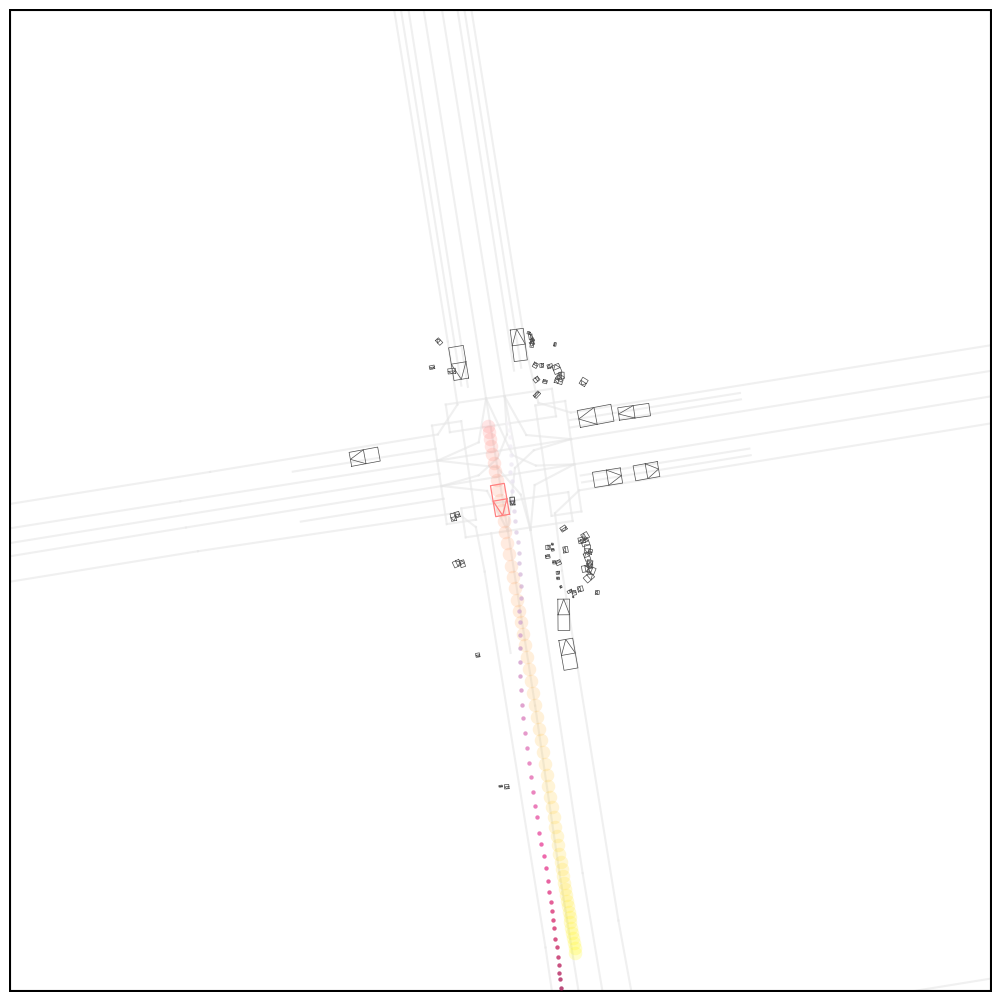}};

\node[right=of cut-in-good, inner sep=0pt] (cut-in-bad) {\includegraphics[width=\imagewidth,height=\imagewidth]{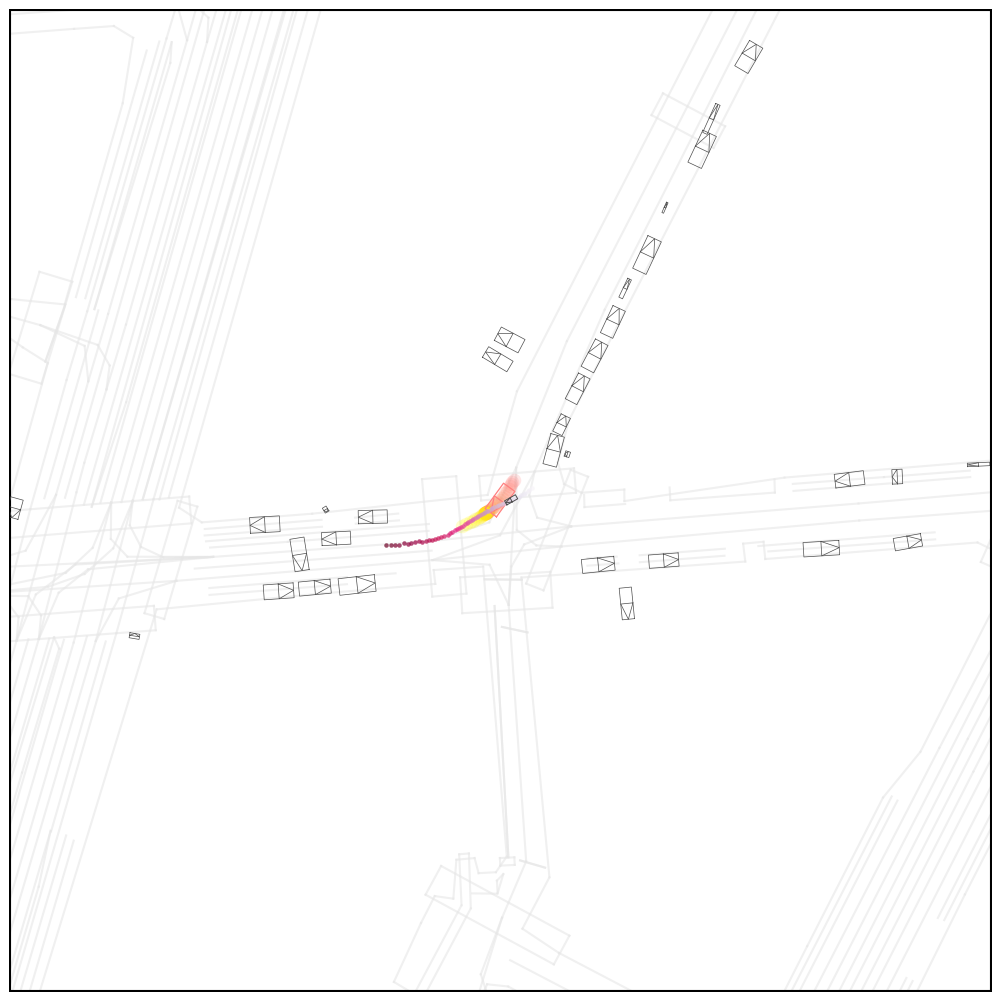}};

\node(tailgate) at (0, -\imagewidth-\rowsep) {\textbf{Tailgate  (left: success, right: failure)}};

\node[below=4pt of tailgate.south west, inner sep=0pt, anchor=north west] (tailgate-good) {\includegraphics[width=\imagewidth,height=\imagewidth]{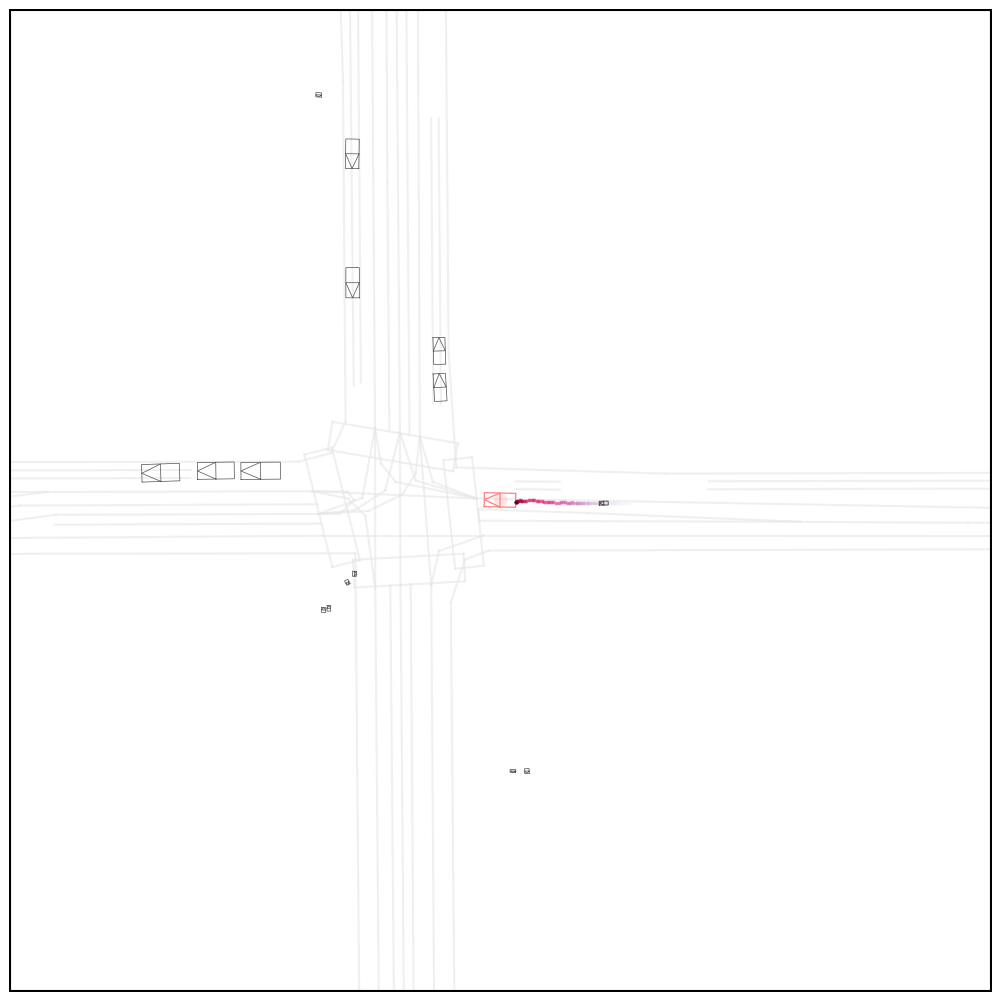}};

\node[right=of tailgate-good, inner sep=0pt] (tailgate-bad) {\includegraphics[width=\imagewidth,height=\imagewidth]{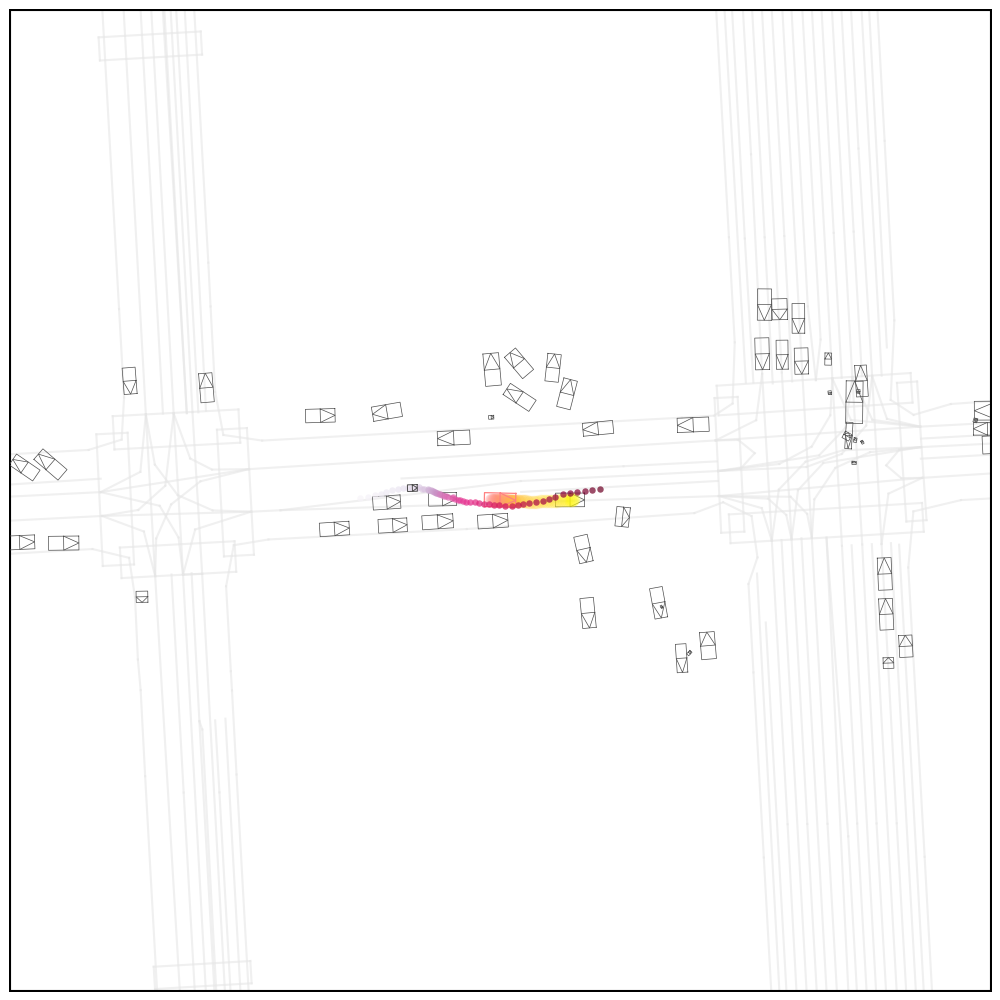}};

\end{tikzpicture}

\begin{tikzpicture}[
    node distance=0.5cm,
    every node/.style={anchor=west},
    box/.style = {draw=#1, line width=0.5mm,inner sep=0.25mm}
]
\def\imagewidth{200pt} %
\def\rowsep{30pt}

\node (s-shape) at (0,0) {\textbf{S-shape  (left: success, right: failure)}};

\node[below=4pt of s-shape.south west, inner sep=0pt, anchor=north west] (s-shape-good) {\includegraphics[width=\imagewidth,height=\imagewidth]{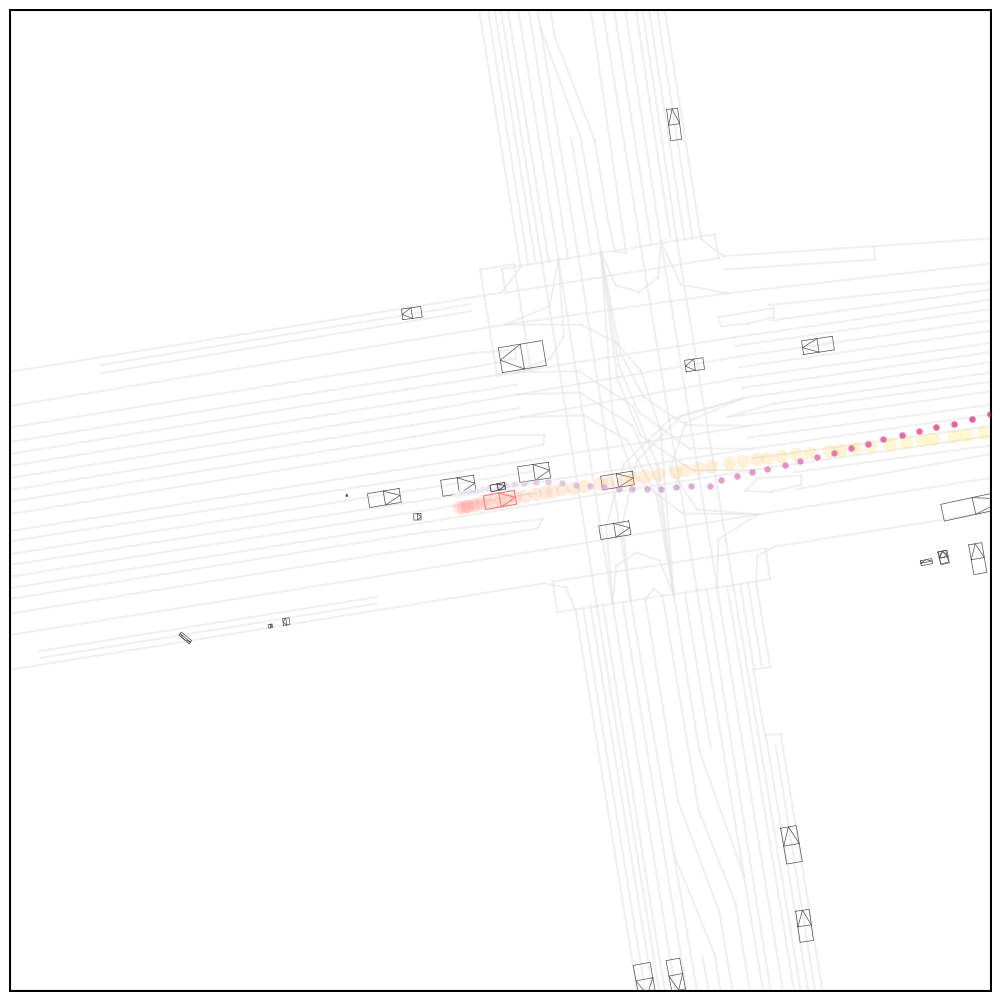}};

\node[right=of s-shape-good, inner sep=0pt] (s-shape-bad) {\includegraphics[width=\imagewidth,height=\imagewidth]{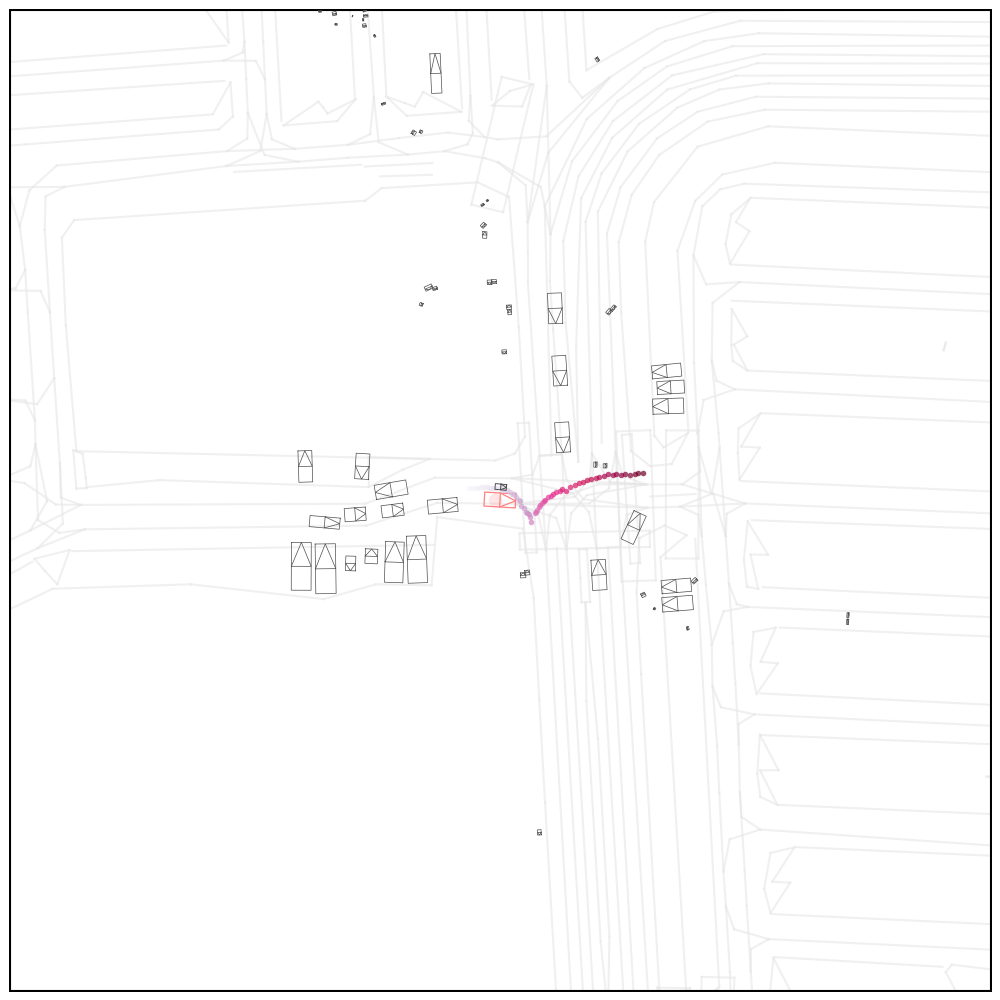}};

\node(surrounding) at (0, -\imagewidth-\rowsep) {\textbf{Surrounding Traffic  (left: success, right: failure)}};

\node[below=4pt of surrounding.south west, inner sep=0pt, anchor=north west] (surrounding-good) {\includegraphics[width=\imagewidth,height=\imagewidth]{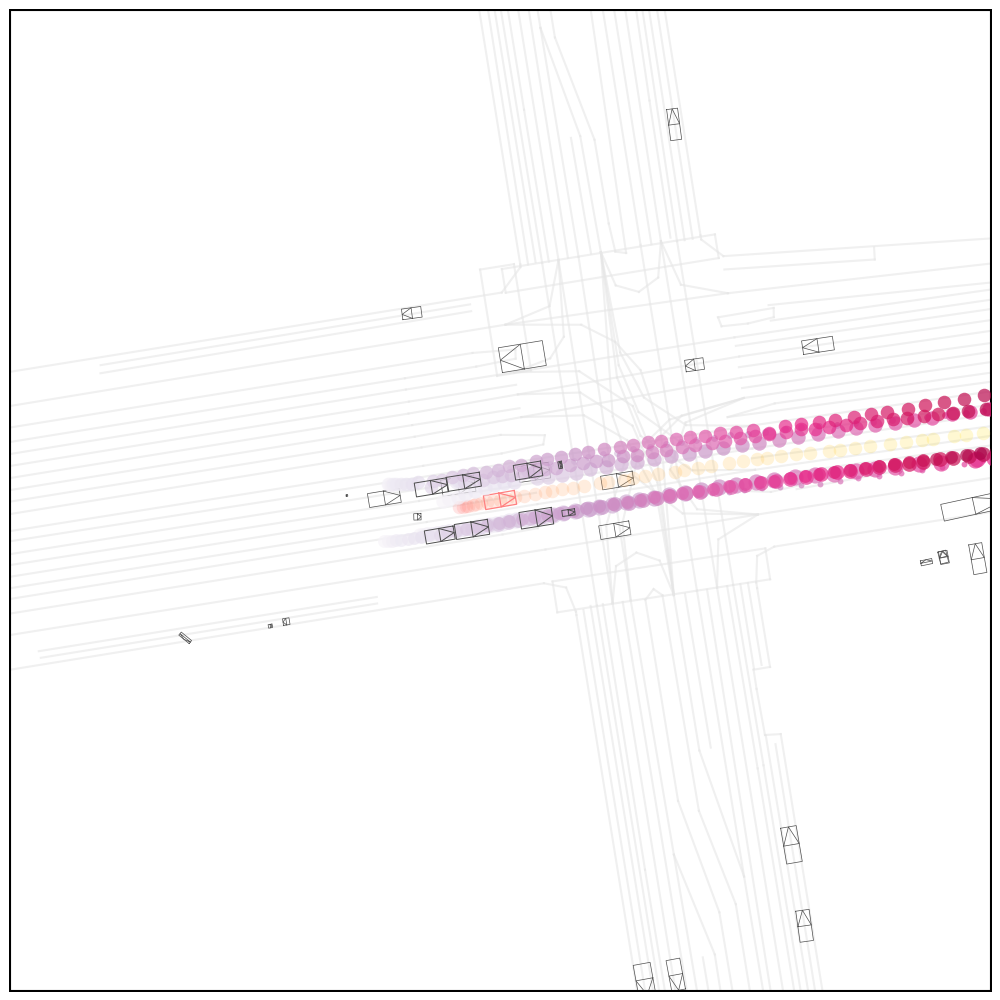}};

\node[right=of surrounding-good, inner sep=0pt] (surrounding-bad) {\includegraphics[width=\imagewidth,height=\imagewidth]{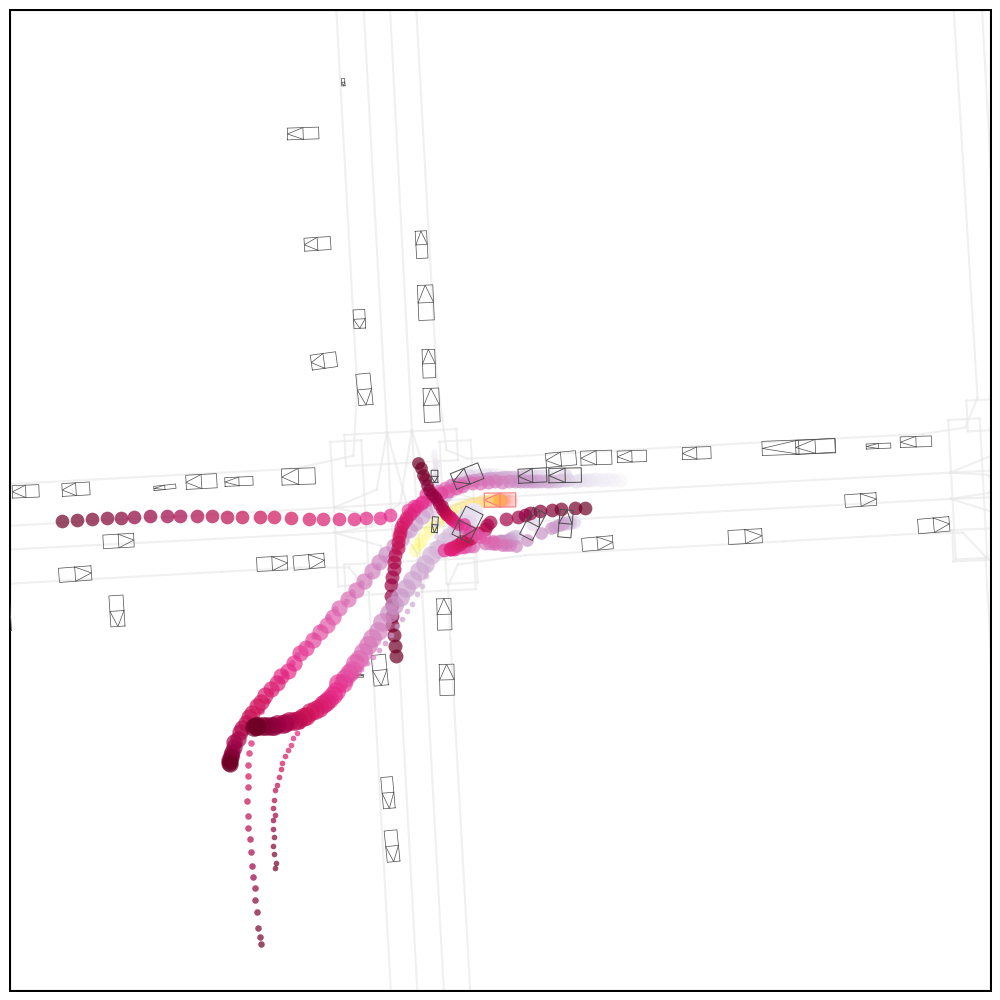}};

\end{tikzpicture}

\subsection{Scene Generation}
\label{subsec:scenegen}

\begin{figure}[ht!]
    \centering
    \includegraphics[trim={1cm 0cm 0cm 0cm },clip,width=\linewidth, keepaspectratio]{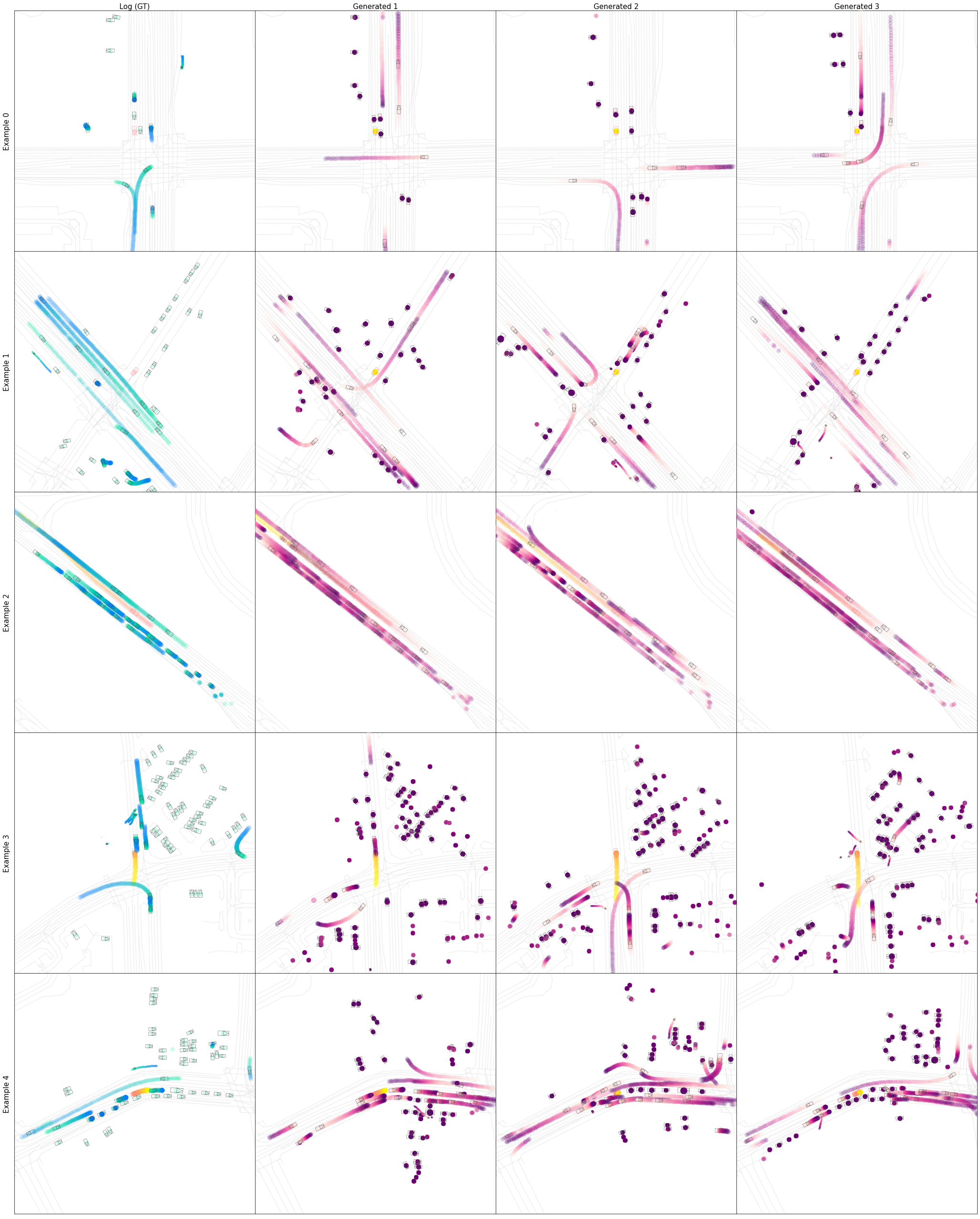}
    \caption{Results of unconditioned scene generation for randomly selected road locations. For each example, we show the ground truth log along with 3 generated scenes.}
    \label{fig:scenegen-grid}
\end{figure}

We show additional unconditioned scene generation results in Fig.~\ref{fig:scenegen-grid}.

\subsection{Generalized Hard Constraint Definitions}
\label{subsec:appendix_constraint_definitions}

\boldparagraph{Non-collision Constraints} ensure the boxes of generated agents do not overlap. We define the potential field of agent $a$ to be a rounded square potential $\phi_{a}(x, y) = \frac{1}{(x-x_a)^4 + (y-y_a)^4 + \epsilon}$ if $||(x-x_a, y-y_a)||_2 < 1.5$ else $0$. We define $\text{clip}_{\text{collision}}(\bm{x})=\argmin_{\bm{x}}\big(\sum_{a\in {A}}\sum_{i=\pm 0.5,j=\pm 0.5}\sum_{a'\in A, a'\neq a}\phi_{a'}(x_a + i w_a, y_a + j l_a)\big)$ that minimizes the potential of each agent's corners against all other agents. $(x, y)$ is defined in the normalized space.

\boldparagraph{Range Constraints} limit a certain feature $\bm{x}_d$ within the range of $d_{min}$ and $d_{max}$. In the context of Scene Generation for example, this can be used to limit the length of a vehicle to an arbitrary range, e.g. between 7-9 meters. We have $\text{clip}_{\text{range}}(\bm{x}_d) = \min(\max(\bm{x}_d, d_{\text{min}}), d_{\text{max}})$.

\boldparagraph{Onroad Constraints} ensure that the bounding boxes of specified generated agents stay on road. We define the offroad potential of road graph polyline $i$ to be $\phi_i(x, y) = (x - x_i)^2 + (y - y_i)^2$ if $W_i(x, y) = 0$ else $0$, where $(x_i, y_i)$ is the closest point on the road graph with respect to $(x, y)$ and $W_i(x,y)$ is the winding number of position $(x,y)$ to polyline $i$, such that we only penalize a trajectory for going offroad. We only consider the closest road graph segment and only consider trajectories that are more than $>20\%$ onroad. We define $\text{clip}_{\text{onroad}}(\bm{x})=\argmin_{\bm{x}} \big( \min_{i \in {RG}} \phi_i(x, y) \big)$.

\end{document}